\pdfoutput=1

\documentclass[11pt]{article}

\usepackage[]{ACL2023}

\usepackage{times}
\usepackage{latexsym}
\usepackage[T1]{fontenc}

\usepackage[utf8]{inputenc}

\usepackage{microtype}

\usepackage{inconsolata}

\usepackage{booktabs}
\usepackage{graphicx}
\usepackage{subfigure}
\usepackage{amsmath}
\usepackage{amssymb}
\usepackage{amsfonts}
\usepackage{algorithmic}
\usepackage{algorithm}
\usepackage{bbm}
\usepackage{multirow}
 \usepackage{listings}
 
\usepackage{booktabs}
\usepackage{afterpage}

\usepackage{xtab}
\usepackage{longtable}

\newcommand{\dtrain}{\mathcal{D}_{\text{train}}}
\newcommand{\xtest}{x_{\text{test}}}

\title{Unified Demonstration Retriever for In-Context Learning}

\author{ Xiaonan Li\textsuperscript{1}\thanks{\ \ Equal Contribution}\ , Kai Lv\textsuperscript{1}$^{*}$, Hang Yan\textsuperscript{1}, Tianyang Lin\textsuperscript{1},\\
\textbf{Zhu Wei\textsuperscript{2}\footnotemark[2]\ , Yuan Ni\textsuperscript{3}, Guotong Xie\textsuperscript{3}, Xiaoling Wang\textsuperscript{2},
Xipeng Qiu\textsuperscript{1}\thanks{\ \ Corresponding Authors}} \\
\textsuperscript{1} Shanghai Key Laboratory of Intelligent Information Processing, Fudan University \\
\textsuperscript{1} School of Computer Science, Fudan University  \\\textsuperscript{2}East China Normal University \textsuperscript{3} Pingan Health Tech\\
\textsuperscript{1}\{lixn20, klv21, hyan19, tylin20, xpqiu\}@fudan.edu.cn, \\
\textsuperscript{2}wzhu@stu.ecnu.edu.cn, xlwang@cs.ecnu.edu.cn \\
\textsuperscript{3}\{niyuan442, xieguotong\}@pingan.com.cn
}

\begin{document}
\maketitle
\begin{abstract}

In-context learning is a new learning paradigm where a language model conditions on a few input-output pairs (demonstrations) and a test input, and directly outputs the prediction. It has been shown highly dependent on the provided demonstrations and thus promotes the research of demonstration retrieval: given a test input, relevant examples are retrieved from the training set to serve as informative demonstrations for in-context learning. While previous works focus on training task-specific retrievers for several tasks separately, these methods are often hard to transfer and scale on various tasks, and separately trained retrievers incur a lot of parameter storage and deployment cost. In this paper, we propose \textbf{U}nified \textbf{D}emonstration \textbf{R}etriever (\textbf{UDR}), a single model to retrieve demonstrations for a wide range of tasks. To train UDR, we cast various tasks' training signals into a unified list-wise ranking formulation by language model's feedback. Then we propose a multi-task list-wise ranking training framework, with an iterative mining strategy to find high-quality candidates, which can help UDR fully incorporate various tasks' signals. 
Experiments on 30+ tasks across 13 task families and multiple data domains show that UDR significantly outperforms baselines. Further analyses show the effectiveness of each proposed component and UDR's strong ability in various scenarios including different LMs (1.3B $\sim$ 175B), unseen datasets, varying demonstration quantities, etc.
\end{abstract}

\section{Introduction}

Large language models have shown an impressive \textit{in-context learning} ability for various Natural Language Processing (NLP) tasks~\citep{gpt3,icl_survey}. 
In-context learning (ICL) is a recent learning paradigm where a language model (LM) learns a task by observing a few input-output pairs (demonstrations) and directly output the prediction of the given test input.
Thus ICL can unify a wide range of NLP tasks through one language model's inference without parameter updates, which makes it a promising alternative to supervised fine-tuning~\citep{bert}. 

\begin{figure}[t]
  \centering
  \includegraphics[width=0.35\textwidth]{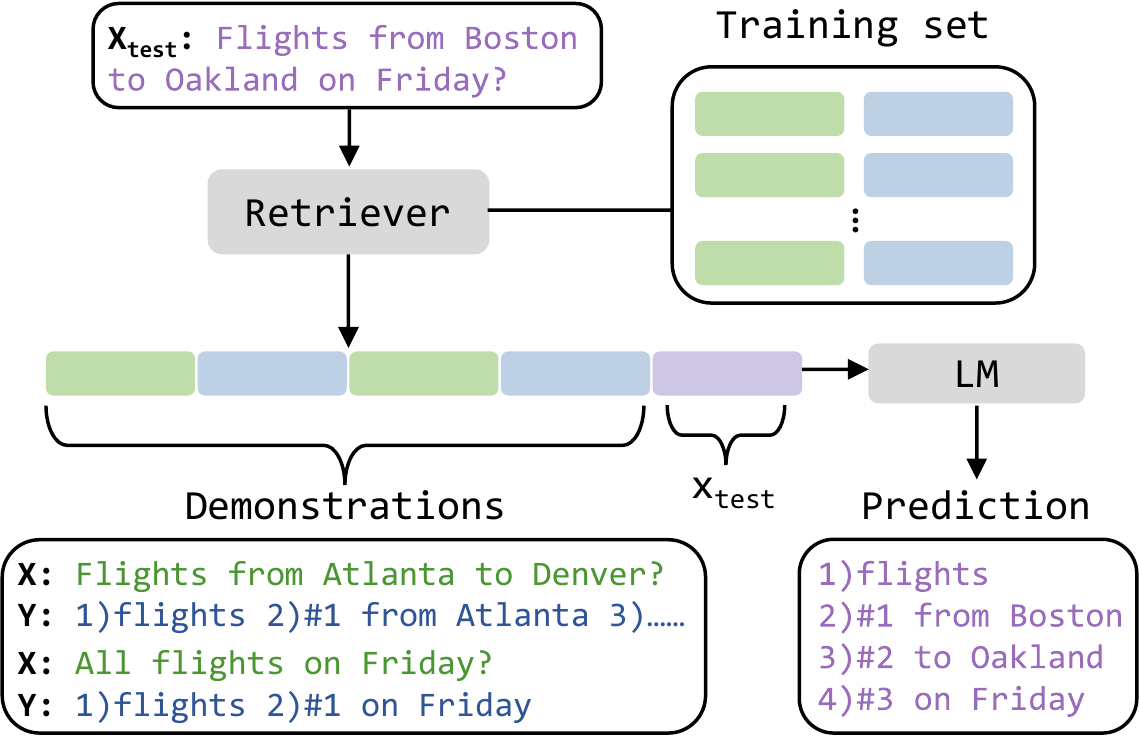}
  \caption{
  Demonstration retrieval: Given a test input $x_{test}$, relevant demonstrations are retrieved from the training set. Then the inference LM takes demonstrations and $x_{test}$ as input and generates the output.
  }
  \vspace{-15pt}
  \label{fig_demonstration_retrieval_first_page}
\end{figure}

However, it has been shown that ICL's performance highly depends on the provided demonstrations~\citep{what_is_good_example_for_gpt3, active_select_for_icl,supporting_examples}. This promotes the research of demonstration retrieval for in-context learning~\citep{what_is_good_example_for_gpt3,epr,cross_lingual_icl_for_text_sql}: As shown in Figure~\ref{fig_demonstration_retrieval_first_page}, given a test input, relevant examples are retrieved from an annotated training set, to serve as informative demonstrations for ICL.

There are about two lines of methods to retrieve demonstrations. One is to leverage off-the-shelf retrievers, e.g., BM25~\citep{bm25} or Sentence-BERT~\citep{sentence_bert}. They can retrieve demonstrations that are textually or semantically similar to the test input and achieve empirical improvements.
Thanks to their versatility, they can serve for extensive NLP tasks, but they are heuristic and sub-optimal since they are not guided by task supervision.
Another line is to train a task-specific retriever by a specially designed task signal. 
\citet{cbr} train the retriever for knowledge-based question answering, based on the logic form's surface similarity.
\citet{ic_dst} explore ICL on dialogue state tracking and design the similarity between dialogue's states as the retriever's training signal.
\citet{epr} and \citet{cross_lingual_icl_for_text_sql} leverage the LM's feedback to train demonstration retrievers for semantic parsing in English and cross-lingual scenarios, respectively. These task-specialized retrievers show better performance than the former,
but they still face two challenges:
1. these explorations are limited to a small range of tasks and demonstrated separately on each task, e.g., semantic parsing or dialogue state tracking, which restricts systematic and compatible research on demonstration retrieval for ICL while ICL is a unified framework for extensive tasks. 
2. it is costly for these methods to transfer and scale on various tasks and the reason is two-fold: (\romannumeral1) they need to design a specialized training signal for each task. (\romannumeral2) the number of retrievers will scale up with increasing tasks, which results in massive parameter storage and deployment costs.

To address these limitations, we explore learning various tasks' demonstration retrieval in a unified formulation and propose \textbf{U}nified \textbf{D}emonstration \textbf{R}etriever (\textbf{UDR}),  a single multi-task model for demonstration retrieval of a wide range of tasks. To train UDR, we cast various tasks' training signals into a unified list-wise ranking formulation. For a training example from task $\mathcal{T}$, we select a list of candidate examples from $\mathcal{T}$'s training set and rank them by LM's feedback. Then we propose a multi-task list-wise ranking training framework, with an iterative mining strategy to find high-quality candidates. Specifically, we iteratively train the retriever to rank candidates and use itself to find high-quality positive candidates and hard negatives. Compared with the representative method for demonstration retrieval, EPR\citep{epr}, which trains the retriever by the binary label from LM's feedback and selects candidates in a manually limited range, our training framework can explore the entire dataset to get high-quality candidates and help UDR fully incorporate the LM's feedback through list-wise ranking training.

Experiments on 30+ tasks across 13 task families and multiple data domains show that UDR significantly outperforms baselines and further analyses show the effectiveness of each proposed component and UDR's strong ability under various scenarios including different LMs (1.3B $\sim$ 175B), unseen datasets, varying demonstrations quantities, etc. We release the code and model checkpoint at \href{https://github.com/KaiLv69/UDR}{https://github.com/KaiLv69/UDR}.

\section{Unified Demonstration Retriever}

\begin{figure*}[t]
  \centering
    \vspace{-10pt}
    \includegraphics[width=0.95\textwidth]{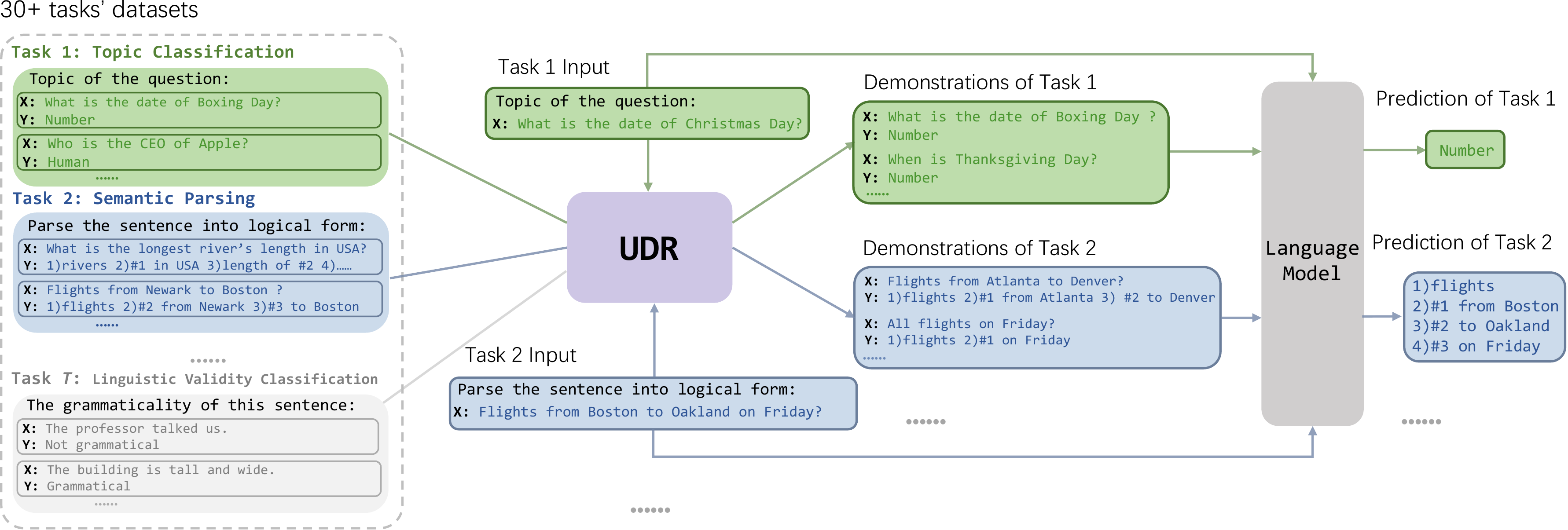}
  \caption{
  Illustration of UDR's inference for various tasks:
  Given a test input and its task's instruction, UDR can retrieve informative demonstrations from the corresponding datasets for ICL, where arrows and lines with various colors such as \includegraphics[width=.4cm]{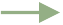} and \includegraphics[width=.4cm]{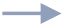} indicate corresponding tasks' pipelines, respectively. 
  } 
  \label{fig_method_details}
  \vspace{-15pt}
\end{figure*}

Provided a language model $G$, a training set $\dtrain$ and a test case $\xtest$, demonstration retrieval aims to retrieve $\xtest$'s relevant demonstrations from $\dtrain$ to help LM $G$ decode the target output. Previous works~\citep{cbr,epr,cross_lingual_icl_for_text_sql} propose task-specialized methods for several tasks separately, but they are hard to transfer and scale on various tasks.
In this work, we focus on learning various tasks' demonstration retrieval in a unified formulation and propose 
UDR
, a single model for demonstration retrieval of a wide range of tasks, as shown in Figure~\ref{fig_method_details}.
We introduce its architecture, training, and inference as follows.

\subsection{Bi-encoder with Task Instruction}
UDR is based on the prevailing bi-encoder architecture, dense passage retriever (DPR)~\citep{dpr}, which encodes the query example and candidate examples separately and then calculates their similarity. 
To distinguish examples from different tasks, UDR encodes the example together with its task instruction, which is a short piece of text related to the task objective. Taking CNN/DailyMail~\citep{data_cnn_dm} as an example, its task instruction can be ``Summarize the text''.
Given an example query $x$ and a candidate demonstration $z=\{x',y'\}$ from task $T_i$, UDR uses the query encoder $E_q$ and demonstration encoder $E_d$ to encode them respectively and calculates their similarity as:
\begin{equation}
    \operatorname{sim}(x,z) = E_{q}(I_{i} \oplus   x)^\top E_{d}(I_{i} \oplus  z),
\end{equation}
where $I_i$ is $T_i$'s task instruction and $\oplus$ is the concatenation operator. $E_q$ and $E_d$ are two multi-layer Transformer~\citep{transformer} encoders with ``CLS'' pooling and can be initialized with pre-trained models~\citep{bert}.

Thus, we can not only get task-specific features by specifying the task instruction, but also retain the uniformity and parameter efficiency of ICL.

\subsection{Learning from LM Feedback}

To train the demonstration retriever, previous works~\citep{cbr,epr,ic_dst} design task-specific training signals for several tasks separately, which makes their methods hard to transfer and scale on various tasks, and hinders systematic and compatible research on demonstration retrieval. 
For UDR's training, we propose to cast various tasks' training signals into a unified list-wise ranking formulation. Then we introduce a multi-task list-wise ranking training framework, 
where we iteratively let the retriever itself to mine high-quality candidates and learn to rank them in turn, across various tasks,
shown in Algorithm~\ref{alg_udr_training}. We introduce the list-wise ranking training and iterative mining strategy as follows.

\subsubsection{Ranking Candidates by LM}
Given a training example $(x,y)$ and its candidates $Z=\{z_i\}_{i=1}^l$, we first rank these candidates as:
\begin{align}
r(z_j) &= rank(s(z_j)|\{s(z_i)\}_{i=1}^l) \\
s_{gen}\left(z_j\right)&=p_G\left(y \mid z_j, x \right), \\
s_{cls}\left(z_j\right)&=\frac{p_G\left(y \mid z_j, x \right)}{\sum_{y'\in Y}p_G\left(y' \mid z_j, x \right)},
\end{align}
where $s(z_j)=s_{gen}(z_j)$ for generation tasks and $s(z_j)=s_{cls}(z_j)$ for classification and multi-choice tasks. $p_G(\cdot|\cdot)$ is the LM $G$'s conditional likelihood. $Y$ is the label space or choices of the classification or multi-choice task, respectively. For simplicity, we omit special tokens and classification tasks' verbalizers in the equations above. 

First we use $G$ to score each candidate~\citep{epr} and calculate $s(z_j)$ as the ground truth $y$'s likelihood conditioned on the candidates $z_j$ and the query input $x$. $s(z_j)$ indicates the importance of $z_j$ for $G$ to encode $x$ and generate the ground truth $y$. Then we rank $Z$ according to $\{s(z_i)\}_{i=1}^l$. The more important $z_j$ is for $x$, the higher $z_j$'s rank will be. Thus we unify various tasks' training signals into the same list-wise ranking formulation using LM's feedback, instead of designing task-specific objectives~\citep{cbr,ic_dst}.

\subsubsection{Loss Function}
With these candidates' ranks from $G$'s feedback, we propose to use the following loss function to inject the ranking signal into the retriever $E$, inspired by LambdaRank~\citep{lambdarank}:
\begin{equation}
\hspace{-5pt}
\mathcal{L}_{rank} = \hspace{-5pt}
\sum\limits_{z_i,z_j\in Z} \hspace{-5pt} w * \log(1+e^{\operatorname{sim}(x,z_j) - \operatorname{sim}(x,z_i)})
\end{equation}
where $w=\operatorname{max}(0,\frac{1}{r(z_i)}-\frac{1}{r(z_j)})$. 

For those $z_i$ and $z_j$ where $r(z_i)$ < $r(z_j)$, $L_{rank}$ will draw $\operatorname{sim}(x,z_i)$ up and optimize the retriever towards $\operatorname{sim}(x,z_i)$ > $\operatorname{sim}(x,z_j)$. Additionally, $w$ adjusts the weight for each pair of demonstrations and inject list-wise ranking information into $\mathcal{L}_{rank}$.
When $z_i$ has a much higher rank than $z_j$, e.g, $r(z_i)=1$ and $r(z_j)=10$, $w$ will be a high weight and strongly draw $\operatorname{sim}(x,z_i)$ up from $\operatorname{sim}(x,z_j)$. Since we optimize the retriever on demonstration pairs under different $w$, $\mathcal{L}_{rank}$ can help UDR fully incorporate candidates' listwise ranking signals from $G$'s feedback for various tasks and learn to retrieve those helpful demonstrations.

To fully leverage the computation of the same batch, we also use the in-batch negative loss as:
\begin{equation}
    \mathcal{L}_{ib} = -\log\frac{e^{\operatorname{sim}(x,z^*)}}{\sum_{z \in \mathbb{Z}}e^{\operatorname{sim}(x,z)}},
\end{equation}
where $z^*$ is the rank-1 candidate of $x$ and $\mathbb{Z}$ is all candidates ($x$'s or not $x$'s) in the batch. 
Each batch is sampled from the same task, and to alleviate the bias towards high-resource tasks, we sample each task according to the multinomial distribution with probabilities $\{p(\mathcal{T}_i)\}_{i=1}^T$ as:
\begin{equation}
    p(\mathcal{T}_i) = \frac{q_i^{\alpha}}{\sum_{j=1}^T q_j^{\alpha}} \ \ \textrm{with} \ \ q_i=\frac{|\mathcal{D}^{\mathcal{T}_i}|}{\sum_{j=1}^T |\mathcal{D}^{\mathcal{T}_j}|},
\end{equation}
where $\mathcal{D}^{\mathcal{T}_i}$ is the $i$th task's dataset. $\alpha$ is a pre-defined hyper-parameter and we follow~\citet{DBLP:conf/nips/ConneauL19} to set $\alpha$ as 0.5.

The overall loss function of UDR is the integration of these two losses as follows, 
\begin{equation}
\mathcal{L} = \lambda * \mathcal{L}_{rank} + (1-\lambda) * \mathcal{L}_{ib},
\end{equation}
where $\lambda$ is a pre-defined hyper-parameter.

\subsubsection{Iterative Candidate Mining}
The selection of candidates can be a key factor for retriever's training~\citep{dpr,ance}. It is desirable for UDR to take the entire training set as candidates to provide abundant ranking signals. However, it is infeasible since scoring all pairs of training examples is quadratic in $|\mathcal{D}|$ and costly. Previous work~\citep{epr} selects those examples which have textually similar targets with $x$’s as candidates. However, it may bias the retriever to learn among candidates with highly similar targets. 
Meanwhile, it can probably miss important demonstrations. For instance, 
if an example $z$ contains relevant logic with the query $x$ but has a dissimilar target with $x$'s, the valuable $z$ will not be selected as candidate to provide signal for the retriever. 
So, we propose an iterative mining strategy to select candidates by the retriever itself. Specifically, we iteratively train the retriever and use it to select candidates in turn. At each iteration, we update each training example's candidates as:
\begin{equation}
    Z^* = \text{top-}K_{z\in \mathcal{D}} \operatorname{sim}(x,z)
\end{equation}
where $\mathcal{D}$ is the task's entire training set. 

Then we will use LM $G$ to score and rank $Z^*$. 
The new candidates in $Z^*$ can be divided into two categories. If a new candidate $z$ has a low score, it means that we find a hard-negative candidate that can provide crucial negative signal for the retriever. If the score of $z$ is high and even higher than all old candidates, it means that we find a valuable positive candidate that can help the retriever learn to find informative demonstrations. 
Thus, with iterative mining, we can explore the entire dataset, find high-quality candidates and improve training progressively.
Before the first iteration, the retriever is untrained, so we initialize candidates based on surface similarity, inspired by \citet{epr}.

For computational efficiency, we first update candidates and score $Z^*$ at each iteration, and then randomly sample $l$ of $Z^*$ and rank them at each training step. In summary, Algorithm~\ref{alg_udr_training} shows the UDR's overall training procedure.
{
\begin{algorithm}[t]
	\caption{Multitask List-wise Ranking Training}
 \label{alg_udr_training}

	\begin{algorithmic}[1]
  	\REQUIRE Bi-encoder $E_q$ and $E_d$, language model $G$, Training sets of $T$ tasks $\{\mathcal{D}^{\mathcal{T}_i}\}_{i=1}^T$
  \STATE Initialize the bi-encoder.
  \STATE Initialize candidates of each training example.
  \STATE Score initialized candidates by $G$.
  
  \FOR{Each iteration}
 
        \FOR{Each training step, $\mathcal{T}_i\sim p(\mathcal{T})$}
        \STATE Sample a batch of examples.
        \STATE For each example, sample $l$ examples $z_{1\sim l}$ from its candidates and rank $z_{1\sim l}$ by $G$'s score.
        \STATE Update the bi-encoder's parameters by $\mathcal{L}$.

		\ENDFOR
  \STATE Update candidates by new $\operatorname{E}_q$ and $\operatorname{E}_d$.
  \STATE Score new candidates by $G$.
  \ENDFOR

	\end{algorithmic} 
\end{algorithm}
}

\subsection{Inference}
After training, we encode each task $\mathcal{T}_i$'s training set using $E_d(p_i\oplus\cdot)$. At the test stage, given a task $\mathcal{T}_i$'s input, $x_{test}$, we use $E_q(p_i\oplus\cdot)$ to compute its encoding and then use FAISS~\citep{faiss} to search over $\mathcal{T}_i$'s training set to find the most relevant demonstrations, ascendingly sorted by $\operatorname{sim}(x_{test},\cdot)$, ${D}=(z_1,z_2,\cdots,z_L)$. 
For generation tasks, the number of final demonstrations, $L$, is determined by the LM $G$'s maximal input length $C$. Specifically, $\sum_{i=1}^L|z_i| + |x_{test}| + |y| \leq C$, where $|y|$ is the pre-defined maximal length of the generated target. 
For classification and multi-choice tasks, we observe that increasing $L$ brings negligible performance improvement and thus we set $L$ to a small value, 8. 
We conduct further analysis of the number of demonstrations in section~\ref{sec_different_number_of_demonstrations}.
Finally, we use greedy decoding to get the result of $G([z_1;z_2;\cdots;z_L;x_{test}])$. Notice that here $D$ is ascendingly sorted by $\operatorname{sim}(x_{test},\cdot)$ unless otherwise specified. Our analysis in section~\ref{sec_different_order} shows that different orderings lead to similar performance. 
Thus we use the same ordering strategy with EPR~\citep{epr} for fair comparison.
\vspace{-5pt}
\section{Experiment}
\vspace{-5pt}

\subsection{Experimental Settings}
\paragraph{Dataset} We train UDR on a wide range of NLP tasks, consisting of about 40 tasks across 13 task families and multiple data domains, including: \textbf{Sentiment Classification}: SST-2, SST-5 \citep{data_sst2_and_sst5}, Amazon \citep{data_amazon}, Yelp \citep{data_yelp_agnews_yahoo}, MR \citep{data_mr} and CR \citep{data_Cr}; \textbf{Topic Classification}: AGNews, Yahoo \citep{data_yelp_agnews_yahoo}, TREC \citep{data_trec} and DBPeida \citep{data_dbpedia}; \textbf{Multi Choice}: COPA \citep{data_copa}, Cosmos QA \citep{data_cosmos}, Commonsense Validation and Explanation (ComE and ComV) \citep{data_come_comv}; \textbf{NLI}: MNLI \citep{data_mnli}, SNLI \citep{data_snli} and RTE \citep{data_rte}; \textbf{Subjectivity Classifcation}: Subj \citep{data_subj}; \textbf{Lingustic Acceptibility}: COLA; \textbf{Semantic Parsing}: BREAK \citep{data_break}, MTOP \citep{data_mtop} and SMCalFlow \citep{data_SMCalFlow}; \textbf{Text Summarization}: CNN/DailyMail \citep{data_cnn_dm}, PubMed \citep{data_pubmed} and Reddit \citep{data_reddit}; \textbf{Commonsense Generation}: CommonGen \citep{data_common_gen}; \textbf{Story Generation}: Roc Story and Ending Generation \citep{data_roc}; \textbf{Code Summarizaton}: Go, Python, Java and PHP \citep{data_codexglue}; \textbf{Text Simplifiction}: WikiAuto + Turk/ASSET \citep{data_wiki_auto}; \textbf{Data to Text}: DART \citep{data_dart} and E2E \citep{data_e2e}.
These tasks' input/output, 
 statistics, split and evaluation metrics are in Appendix~\ref{appendix_task_overview}.

\begin{table*}[t]
\small
\centering
\setlength{\tabcolsep}{2.7 mm}
\begin{tabular}{@{}lcccccccccc@{}}
\toprule
\multirow{2}{*}{\textbf{\begin{tabular}[c]{@{}l@{}}Retrieval \\ Method\end{tabular}}} & \multicolumn{6}{c}{\textbf{Sentiment Classification}}                                         & \multicolumn{4}{c}{\textbf{Topic Classification}}             \\
\cmidrule(lr){2-7}
\cmidrule(lr){8-11}
& SST-2         & SST-5         & Amazon        & Yelp          & MR            & CR            & AGNews        & TREC          & DBPedia       & Yahoo         \\ \midrule
Random                                                                                & 57.7          & 28.2          & 23.9          & 25.3          & 56.0          & 52.4          & 74.2          & 42.6          & 73.7          & 39.1          \\
BM25                                                                                  & 74.1          & 38.3          & 31.6          & 36.9          & 71.4          & 57.2          & 88.4          & 89.4          & 97.2          & 62.5          \\
SBERT                                                                                 & 84.3          & 40.0          & 33.4          & 36.0          & 79.0          & 61.3          & 88.3          & 89.4          & 96.7          & 58.4          \\
Instructor                                                                            & 83.7          & 42.4          & 42.4          & 46.6          & 78.5          & 64.1          & 89.6          & 91.2          & 97.7          & 67.2          \\
EPR                                                                                   & 87.9          & 46.9          & 49.1          & 49.6          & 80.6          & 65.7          & 89.9          & 95.2          & 98.1          & 66.1          \\
UDR                                                                                   & \textbf{92.4} & \textbf{50.5} & \textbf{54.9} & \textbf{61.7} & \textbf{85.2} & \textbf{82.6} & \textbf{91.5} & \textbf{96.6} & \textbf{98.7} & \textbf{67.5} \\ \bottomrule
\end{tabular}

\setlength{\tabcolsep}{2.4 mm}

\vspace{2pt}
\begin{tabular}{@{}lcccccccccc@{}}
\multirow{2}{*}{\textbf{\begin{tabular}[c]{@{}l@{}}Retrieval \\ Method\end{tabular}}} & \multicolumn{4}{c}{\textbf{Multi Choice}}                     & \multicolumn{3}{c}{\textbf{NLI}}              & \multicolumn{2}{c}{\textbf{Other}} & \multicolumn{1}{l}{\multirow{2}{*}{\textbf{Overall}}} \\ 
\cmidrule(lr){2-5}
\cmidrule(lr){6-8}
\cmidrule(lr){9-10}
& COPA          & Cosmos QA     & ComE          & ComV          & MNLI          & SNLI          & RTE           & Subj             & COLA            & \multicolumn{1}{l}{}                                  \\ \midrule
Random                                                                                & 71.6          & 26.2          & 41.4          & 50.5          & 34.1          & 33.0          & 55.6          & 60.0             & 52.8            & 47.3                                                  \\
BM25                                                                                  & 71.2          & 27.1          & 41.4          & 50.9          & 35.3          & 41.5          & 50.5          & 78.8             & 53.3            & 57.7                                                  \\
SBERT                                                                                 & 72.4          & 27.3          & 41.1          & 50.3          & 38.0          & 42.0          & 49.8          & 88.7             & 56.3            & 61.6                                                  \\
Instructor                                                                            & 71.6          & 27.1          & 41.9          & 49.9          & 41.3          & 46.7          & 52.7          & 84.3             & 56.0            & 63.2                                                  \\
EPR                                                                                   & \textbf{73.2} & 28.4          & 43.0          & 50.4          & 54.3          & 74.0          & 55.6          & 92.1             & 70.3            & 68.8                                                  \\
UDR                                                                                   & {72.8} & \textbf{29.9} & \textbf{45.6} & \textbf{63.9} & \textbf{73.8} & \textbf{83.6} & \textbf{65.3} & \textbf{95.0}    & \textbf{78.9}   & \textbf{73.2}                                         \\ \bottomrule
\end{tabular}
\vspace{-5pt}
\caption{Main results on classification and multi-choice tasks.}
\vspace{-5pt}
\label{tab_main_result_cls}
\end{table*}

\begin{table*}[t]
\small
\centering
\setlength{\tabcolsep}{1.25 mm}

\begin{tabular}{@{}lccccccccc@{}}
\toprule
\multirow{2}{*}{\textbf{\begin{tabular}[c]{@{}l@{}}Retrieval \\ Method\end{tabular}}} & \multicolumn{3}{c}{\textbf{Semantic Parsing}} & \multicolumn{3}{c}{\textbf{Text Summarizaiton}} & \textbf{CommonGen} & \multicolumn{2}{c}{\textbf{Story Generation}} \\ 
\cmidrule(l){2-4} 
\cmidrule(l){5-7} 
\cmidrule(l){8-8} 
\cmidrule(l){9-10} 
& BREAK         & MTOP          & SMCalFlow     & CNN/DM         & PubMed         & Reddit        & CommonGen          & Roc Story             & Roc Ending            \\ \midrule
Random                                                                                & 1.9           & 6.6           & 8.7           & 20.8           & 23.6           & 15.6          & 21.1               & 9.3                   & 13.4                  \\
BM25                                                                                  & 26.0            & 52.9          & 46.1          & 18.6           & 24.5           & 15.3          & 26.0               & 12.3                  & 19.2                  \\
SBERT                                                                                 & 22.4          & 48.6          & 43.1          & 19.2           & 25.2           & 15.4          & 25.7               & 12.2                  & 19.1                  \\
Instructor                                                                            & 22.7          & 50.5          & 46.3          & 19.0           & 24.8           & 15.3          & 26.5               & 12.4                  & 21.8                  \\
DR-Target                                                                             & 22.1          & 49.6          & 41.6          & 19.4          & 24.6           & 16.0          & 24.5              & 11.9                  & 20.1                  \\
EPR                                                                                   & 31.9          & 64.4          & 54.3          & 20.3           & 24.8           & 15.5          & 25.3               & 12.9                  & 21.2                  \\
UDR                                                                                   & \textbf{35.2} & \textbf{66.8} & \textbf{60.4} & \textbf{21.2}  & \textbf{26.1}  & \textbf{16.2} & \textbf{27.1}      & \textbf{17.6}         & \textbf{24.7}         \\ \bottomrule
\end{tabular}
\vspace{2pt}
\setlength{\tabcolsep}{2.85 mm}

\begin{tabular}{@{}lcccccccccc@{}}
\multirow{2}{*}{\textbf{\begin{tabular}[c]{@{}l@{}}Retrieval \\ Method\end{tabular}}} & \multicolumn{4}{c}{\textbf{Code Summarization}}               & \multicolumn{3}{c}{\textbf{Text Simpilification}} & \multicolumn{2}{c}{\textbf{Data to Text}} & \multirow{2}{*}{\textbf{Overall}} \\ 
\cmidrule(l){2-5} 
\cmidrule(l){6-8} 
\cmidrule(l){9-10} 
& Go            & Python        & Java          & PHP           & WikiAuto        & Turk           & ASSET          & DART                & E2E                 &                                   \\ \midrule
Random                                                                                & 27.3          & 7.9           & 6.7           & 18.9          & 8.3             & 28.0           & 24.8           & 20.4                & 21.9                & 15.8                              \\
BM25                                                                                  & 30.4          & 9.7           & 11.7          & 23.6          & 10.2            & 29.1           & 26.6           & 28.4                & 29.2                & 24.2                              \\
SBERT                                                                                 & 28.3          & 13.7          & 15.1          & 22.0          & 9.5             & 29.1           & 26.7           & 27.9                & 24.2                & 23.7                              \\
Instructor                                                                            & 29.9          & 11.5          & 13.1          & 24.0          & 11.3            & 29.0           & 26.3           & 28.7                & 22.4                & 24.2                              \\
DR-Target                                                                          & 28.1          & 12.2          & 13.0          & 24.2          & 10.8            & 29.4           & 26.7           & 30.1                & 24.7                & 23.8                              \\
EPR                                                                                   & \textbf{30.5}          & 17.4          & 17.4          & 30.2          & 13.3            & 30.8           & 27.6           & 31.8                & 29.3                & 27.7                              \\
UDR                                                                                   & 29.4 & \textbf{22.3} & \textbf{25.2} & \textbf{33.2} & \textbf{19.5}   & \textbf{32.9}  & \textbf{32.1}  & \textbf{34.5}       & \textbf{32.6}       & \textbf{30.9}                     \\ \bottomrule
\end{tabular}
\vspace{-4pt}
\caption{Main results on generation tasks. }
\vspace{-15pt}
\label{tab_main_result_gen}

\end{table*}

\paragraph{Implementation Details}
We follow EPR~\citep{epr} to use GPT-Neo-2.7B~\citep{gpt_neo} as the scoring LM and the inference LM for most experiments in the paper unless otherwise specified. We also explore UDR's transferability across different inference LMs in section~\ref{sec_transferability_different_lm}. Following EPR\citep{epr}, we initialize $E_q$ and $E_d$ as two separate ``BERT-base-uncased'' encoders~\citep{bert}. We list the overall hyper-parameters and implementation details in Appendix~\ref{appendix_implementation_details_and_hyper_parameters}. On each task, we use one specific template for scoring and inference (see Appendix~\ref{appendix_task_overview}). 
We evaluate UDR's performance when inference templates are different with the scoring template in Appendix~\ref{appendix_different_inference_template}, and the results show that UDR has stable performance across varying inference templates, which reflects UDR's generality.

\paragraph{Model Comparison}
With the same inference LM, GPT-Neo-2.7B, we compare UDR with previous methods for demonstration retrieval by the downstream ICL performance, including:
\textbf{1. Random}: We randomly sample demonstrations from the corresponding task's training set.
\textbf{2. BM25}~\citep{bm25}: A prevailing sparse retriever. For each test input $x_{test}$, we use BM25 to retrieve examples with the most similar input.
\textbf{3. SBERT}~\citep{sentencebert}: We use the Sentence-BERT as the dense demonstration retriever. Specifically, we follow ~\citet{epr} to take ``paraphrase-mpnet-base-v2'' to encode the test input $x_{test}$ and training set's inputs, and retrieve the examples with the most similar input as demonstrations.
\textbf{4. Instructor}~\citep{instructor}: Instructor is a recently proposed competitive text embedding model trained on 330 tasks with instructions. By providing  the specialized instruction, it can serve for demonstration retrieval. 
For fair comparison, we conduct experiments on its released base-size model.
\textbf{5. DR-Target}: This baseline is inspired by previous works on generation tasks like dialogue state tracking, question answering and code generation~\citep{ic_dst,cbr,reliable_code_generation}, which design the task-specific target's similarity and use examples with similar targets to train the retriever. Here we use BM25 as the similarity function for each task's target output. Specifically, we use BM25 to find positive pairs with similar targets and use DPR~\citep{dpr} for training.
\textbf{6. EPR}~\citep{epr}: EPR is a recently proposed representative method for training demonstration retriever. It uses the language model to assign candidate examples with positive and negative labels and thus trains a task-specific demonstration retriever by DPR. For fair comparison, we train EPR on each task using the same hyper-parameters of UDR. Specially, we discuss EPR's candidate quantity in Appendix~\ref{appendix_implementation_details_and_hyper_parameters}.

Except that the performance of Random, BM25, SBERT and EPR on semantic parsing is from the previous paper~\citep{epr}, other results are from our implementation since they are not explored previously.

\subsection{Main Results}
We show the performance comparison of classification tasks and generation tasks in Table~\ref{tab_main_result_cls} and Table~\ref{tab_main_result_gen}, respectively. We can see that UDR outperforms baselines significantly on most tasks, which shows UDR's best overall demonstration retrieval ability on a wide range of NLP tasks. 
Specially, compared with DR-Target and EPR, UDR has better overall performance and this shows the effectiveness of our unification of various tasks' training signals.
Meanwhile, compared with Instructor~\citep{instructor}, the text embedding model trained on 330 tasks' text pairs, UDR has an improvement of 10 and 6.7 points for classification and generation tasks respectively with less training data. This straightly demonstrates that our proposed training framework can help UDR incorporate LM's feedback through a unified ranking formulation and better retrieve informative demonstrations. 

Additionally, we find the random baseline shows the worst performance on most tasks and this reflects the necessity to retrieve high-quality relevant demonstrations. Meanwhile, EPR and UDR have better performance than other methods, which reflects the importance of LM's feedback. Among these datasets, we notice a different trend on text summarization datasets like CNN/DailyMail and Reddit, on which these methods have similar performance.
We conjecture that the LM can already have the knowledge of summarization since there are a lot of ``[Article, TL;DR, Abstract]'' texts in its pre-training corpus \citep{gpt2}, thus random demonstrations can well activate LM's summarization ability without example-specific information. 

\subsection{Analysis}
\subsubsection{Ablation Study}
To evaluate the effect of UDR's each component, we conduct ablation study on SMCalFlow, SST-2 and Java code summarization, shown in Table~\ref{tab_ablation}. 
When removing list-wise ranking training, we use EPR's training strategy~\citep{epr}.
We can see that removing task instructions cause slight performance degradation, which indicates that they can help UDR distinguish examples from various tasks and thus get better task-specific features.
Meanwhile, we can see that UDR has a slightly better performance than the single-task counterpart on SST-2 and Java. We suppose that is because there are several relevant tasks in UDR's training tasks and our multi-task ranking unification can help UDR fully share these tasks' knowledge. The performance of single-task UDR still outperforms EPR significantly and this straightly reflects that our training components, i.e., list-wise ranking formulation and iterative candidate mining strategy, can 1. help UDR better incorporate LM's feedback than EPR 2. serve as a competitive universal training method for a task-specific retriever. Removing list-wise ranking training and iterative candidate mining both cause performance degradation, which straightly indicates their effectiveness.
\begin{table}[]
\small
\centering
\begin{tabular}{@{}lcccl@{}}
\toprule
                  & \hspace{-15pt}SMCalFlow & SST-2 & Java & Avg  \\ \midrule
UDR               & 60.8      & 91.3  & 23.2 & 58.4 \\
- w/o Task Prompt & 60.1      & 90.8  & 21.9 & 57.6 \\
- w/o MultiTask   & 60.9      & 91    & 22.9 & 58.3 \\
- w/o Rank Loss   & 56.7      & 89.2  & 21.1 & 55.7 \\
- w/o Self-Guided & 59.5      & 90.2  & 19.7 & 56.5 \\ \bottomrule
\end{tabular}
\caption{Ablation study of UDR's each component.}
\label{tab_ablation}
\end{table}

\begin{table}[]
\centering
\small
\setlength{\tabcolsep}{0.6 mm}
\begin{tabular}{@{}lcccccc@{}}
\toprule
Dataset          & \multicolumn{3}{c}{SMCalFlow}           & \multicolumn{3}{c}{E2E} \\ \midrule
LMs / Methods    & BM25 & EPR  & \multicolumn{1}{c|}{UDR}  & BM25   & EPR    & UDR   \\ \midrule
Text-Davinci-003 & 55.0 & 58.9 & \multicolumn{1}{c|}{64.7} & 31.3   & 31.5   & 34.3  \\
Code-Davinci-002 & 50.9 & 55.2 & \multicolumn{1}{c|}{62.9} & 23.5   & 24.4   & 26.4  \\
GPT-J            & 49.0 & 55.9 & \multicolumn{1}{c|}{64.0} & 33.3   & 33.7   & 35.0  \\
GPT-Neo-1.3B     & 44.8 & 52.9 & \multicolumn{1}{c|}{59.5} & 29.9   & 29.7   & 31.9  \\ \midrule
GPT-Neo-2.7B     & 46.5 & 53.7 & \multicolumn{1}{c|}{62.2} & 29.2   & 29.1   & 32.6  \\ \bottomrule
\end{tabular}
\caption{Results on 1000 randomly sampled test examples across different inference LMs.}
\vspace{-5pt}
\label{tab_different_lm}
\end{table}

\subsubsection{Transferability across Different LMs}
\label{sec_transferability_different_lm}
In this section, we evaluate UDR's transferability across different inference LMs on SMCalFlow and E2E. Specifically, we compare BM25, EPR and UDR on inference LMs with different sizes, including: GPT-Neo-1.3B~\citep{gpt_neo}, GPT-J (6B)~\citep{gpt_j}, Code-Davinci-002 (175B) \citep{codex} and Text-Davinci-003 (175B) \citep{gpt3,instruct_gpt} and we show the result in Table~\ref{tab_different_lm}.
When comparing UDR with baselines, the trends are similar with using GPT-Neo-2.7B (the scoring LM) as inference LM. UDR outperforms BM25 and EPR significantly and it shows UDR's strong transferability across different inference LMs. 
Meanwhile, we find that UDR with larger inference LM can improve performance such as Text-Davinci-003 on SMCalFlow and GPT-J on E2E, which shows UDR's potential utility in the future where more competitive large-scale LM is built. When we demonstrate the example-specific demonstration transferability across different inference LMs in this paper, \citet{supporting_examples} show that task-level demonstrations also exhibit such transferability. We leave the analysis of the transferablity of ICL's demonstrations across different LMs as future work.

\subsubsection{Performance on Unseen Datasets}
In this section we explore UDR's zero-shot transferability and evaluate it on unseen datasets including: 1. Twitter sentiment classification~\citep{data_twitter_sentiment} 2. question-answering NLI (QNLI)~\citep{data_glue} 3. Ruby and JavaScript code summarization~\citep{data_codexglue}. These domains or programming languages (Twitter, NLI on QA, Ruby and Javascript) are never seen during UDR's training and thus can straightly reflect UDR's zero-shot transferability. We compare UDR with two powerful universal retrievers, BM25 and SBERT, and show the result in Table~\ref{tab_zero_shot_transfer}. We can see UDR significantly outperforms BM25 and SBERT on these unseen datasets by about 10 points on average, which shows that the learned ranking knowledge inside UDR can be well transferred and generalized to unseen datasets.

\begin{table}[]
\small
\centering
\begin{tabular}{@{}lcccc@{}}
\toprule
      & Twitter & QNLI & Ruby & JavaScript \\ \midrule
BM25  & 50.0    & 54.1 & 9.2  & 12.7       \\
SBERT & 51.6    & 53.7 & 8.7  & 15.9       \\
UDR   & 56.8    & 74.4 & 19.6 & 21.6       \\ \bottomrule
\end{tabular}
\caption{The performance of UDR on unseen datasets.}
\label{tab_zero_shot_transfer}
\end{table}

\subsubsection{The Order of Demonstrations}
\label{sec_different_order}
Previous work~\citep{icl_ordering} has revealed that ICL is sensitive to demonstrations' order when using random examples.
Specifically, the same randomly sampled demonstrations with different orders can lead to the performance between random guess and near state-of-the-art. 
Here we explore the effect of ordering on example-specific demonstrations retrieved by UDR. We compare 3 demonstrations' orders: 1. random, for this setting, we run experiments with 10 different random seeds and report the best and worst performance. 2. descending sorted by UDR's score, i.e, the demonstration which has the highest similarity with $x_{test}$ is put at the beginning of LM's input. 3. ascending sorted by UDR's score, opposite to ``2''. The result is shown in Table~\ref{tab_order}. We observe a different phenomenon from that in previous work~\citep{icl_ordering}. In general, The performance of UDR's demonstrations with different orders is more stable than previously investigated random examples. Across these tasks, different orders' performance gap is within 1 point, and it is far less than the performance fluctuation of up to tens points when using random examples~\citep{icl_ordering}. This indicates that high-quality demonstrations are less sensitive to the ordering and stabilize in-context learning, which is consistent with the analysis in previous work~\citep{sensitivity_and_accuracy,supporting_examples}.

\begin{table}[]
\small
\centering
\setlength{\tabcolsep}{1 mm}
\begin{tabular}{@{}lcccc@{}}
\toprule
                 & SST-2 & TREC & Reddit & CommonGen \\ \midrule
Random-Order$_{\text{Best}}$     & 92.5  & 96.6 & 16.8   & 27.5      \\
Random-Order$_{\text{Worst}}$     & 92.0  & 96.2 & 16.2   & 26.6      \\
Descending-Order & 92.2  & 96.6 & 16.2   & 27.0      \\
Ascending-Order  & 92.4  & 96.6 & 16.3   & 27.3      \\ \bottomrule
\end{tabular}
\caption{The effect of different demonstration orders.}
\vspace{-15pt}
\label{tab_order}
\end{table}
\subsubsection{The Impact of Demonstration Quantity}
\label{sec_different_number_of_demonstrations}
\begin{figure}[t]
\centering
\includegraphics[width=0.45\textwidth]{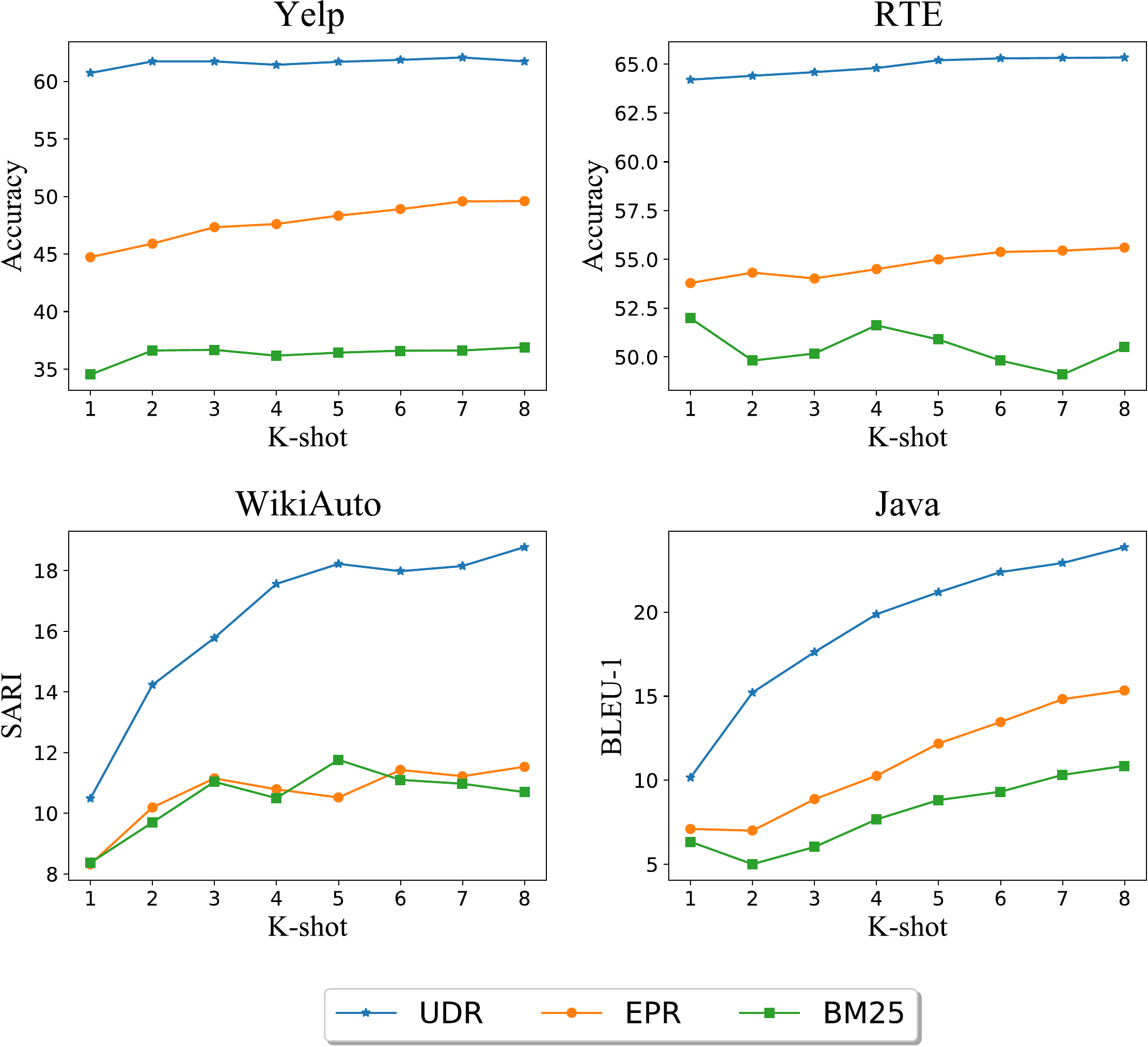}
\caption{The effect of demonstration quantity.}
\vspace{-12pt}
\label{fig_num}
\end{figure}

We compare UDR with BM25 and EPR under different amounts of demonstrations on two classification tasks: Yelp and RTE, and two generation tasks: WikiAuto and Java code summarization. We show results in Figure~\ref{fig_num}. We can see that UDR outperforms baselines consistently across varying amounts of demonstrations. Meanwhile, we can draw two conclusions from the results: 1. The number of demonstrations has a greater impact on generation tasks than classification tasks. Specifically, as the number of demonstrations increases, generation tasks' performance gets significant improvements while classification tasks' has slight or no improvements. 2. The quality of demonstrations can be more important than their quantity. In detail, UDR with the quota of 2 demonstrations still outperforms BM25 and EPR with 8 demonstrations. This also reflects the strong demonstration retrieval ability of UDR. 
\citet{mot} observe the similar trends in the CoT-retrieval scenario, indicating that the relevance of the used reasoning paths is more important than their quantity.

\vspace{-5pt}

\section{Related Work}
\vspace{-5pt}
In this section, we introduce previous demonstration retrievers for in-context learning, and explain the difference between UDR and them. In general, there are two kinds of demonstration retrievers for ICL. One is to leverage off-the-shelf retrievers. For example, ~\citet{what_is_good_example_for_gpt3} propose to use a fine-tuned BERT to encode examples and use a KNN-based method to retrieve semantically similar demonstrations to improve ICL. ~\citet{icl_mt} use BM25 to retrieve demonstrations for machine translation. 
Compared with them, UDR incorporates various tasks' supervision by unified LM's feedback and thus can better retrieve informative demonstrations. 
Another approach is to train a task-specific retriever by a designed task-specific signal. \citet{cbr} explore demonstration retrieval for knowledge-based question answering and define the F1 score of logic forms as soft-label to train the retriever. \citet{reliable_code_generation} train a demonstration retriever for code generation, based on the edit distance of abstract syntax trees. \citet{ic_dst} define the similarity between dialogue states, and use it to train a demonstration retriever for dialogue state tracking. \citet{epr} propose Efficient Prompt Retriever (EPR) for semantic parsing, which is to use the language model to score examples, assign positive and negative labels for them and use DPR~\citep{dpr} to train a demonstration retriever. \citet{cross_lingual_icl_for_text_sql} explore demonstration retrieval for cross-lingual semantic parsing using a similar example scoring method with EPR. 
These task-specific methods serve for each task separately and are hard to transfer and scale on various tasks. For other tasks, it requires to redesign the similarity function or training signal. Compared with them, we introduce a unified training framework based on list-wise ranking and propose a single multi-task retriever UDR to serve for a wide range of tasks. Compared with EPR, besides UDR's versatility on various tasks, UDR can incorporate LM's feedback by ranking-based training in a more fine-grained way and receive more crucial candidates' signals by the iterative mining strategy. 
\citet{sentence_embedding_by_gpt3} propose CLAIF to enhance the sentence embedder by the gigantic language model's feedback. Specifically, they use GPT-3~\citep{gpt3} to generate the data of sentence pairs and then score them by the output of GPT-3, which depends on the strong natural language understanding ability of GPT-3. 
Different from them, we leverage the conditional probability to measure the helpfulness of an example, which only needs a small language model, and is more efficient and environmental-friendly.
Recently, \citet{mot} propose MoT (Memory-of-Thought) to let the LLM self-improve in two stages:
1. Before test stage, the LLM generate reasoning paths and answers on an unlabeled dataset for itself, 
2. At test stage, the LLM retrieves relevant reasoning paths (memory) to help itself answer the given test question. 
While MoT focuses on the scenario with unlabeled dataset and uses the LLM for retrieval, we train a small retriever by a LM's feedback from tasks' supervision and thus the proposed method is more lightweight. We leave demonstration retrieval with reasoning paths or unlabeled datasets as future work.

\vspace{-2pt}
\section{Conclusion}
\vspace{-2.5pt}
In this paper, we propose UDR, a single multi-task model for a wide range of tasks' demonstration retrieval. To train UDR, we cast various tasks' training into a unified list-wise ranking formulation by language model's feedback, and propose a multi-task list-wise ranking training framework, with an iterative mining strategy to find high-quality candidates. Experiments on 30+ tasks show that UDR significantly outperforms baselines. 
Further analyses show the effectiveness of each proposed component and UDR's strong ability in various scenarios including different LMs (1.3B $\sim$ 175B), unseen datasets, varying demonstration quantities, etc.
\section*{Limitations}

We illustrate this paper's limitations from the following three aspects:

1) Limited by the computational resources, we only train UDR from the initialization of ``BERT base uncased'' following EPR~\citep{epr}. We regard explorations based on other competitive pre-trained models like RoBERTa~\citep{roberta} and DeBERTa~\citep{deberta} as future work.

2) Most of current dense demonstration retrievers, including UDR, are black-box models. Although they lead to significantly better performance than BM25, how they find informative demonstrations is still unknown. Therefore, a better understanding of the principle of informative demonstration's retrieval or an interpretable and transparent demonstration retriever may be the next stage of improving demonstration retrieval. \citet{knn_prompting} propose a more explainable method, beyond-context learning, which first uses the language model to
get training data's next word probability distribution, then assigns test instances with labels of their nearest neighbors with similar next word's probability distribution. We leave demonstration retrieval with better explainability as future work.

3) 
In the training stage we use LM to score candidates separately but in the inference stage LM is provided with a sequence of demonstrations.
Although experimental results demonstrate UDR's effectiveness, we think it is a promising direction to model the dependence between different demonstrations and leave it to future work.

\section{Acknowledgements}
This work was supported by the National Natural Science Foundation of China (No. 62236004 and No. 62022027) and Shenzhen City's Science and Technology Plan Project (No. JSGG20210802153806021).
\bibliography{anthology,custom}

\begin{thebibliography}{65}
\expandafter\ifx\csname natexlab\endcsname\relax\def\natexlab#1{#1}\fi

\bibitem[{Agrawal et~al.(2022)Agrawal, Zhou, Lewis, Zettlemoyer, and
  Ghazvininejad}]{icl_mt}
Sweta Agrawal, Chunting Zhou, Mike Lewis, Luke Zettlemoyer, and Marjan
  Ghazvininejad. 2022.
\newblock \href {https://doi.org/10.48550/arXiv.2212.02437} {In-context
  examples selection for machine translation}.
\newblock \emph{CoRR}, abs/2212.02437.

\bibitem[{Amplayo et~al.(2022)Amplayo, Brazinskas, Suhara, Wang, and
  Liu}]{data_Cr}
Reinald~Kim Amplayo, Arthur Brazinskas, Yoshi Suhara, Xiaolan Wang, and Bing
  Liu. 2022.
\newblock \href {https://doi.org/10.1145/3477495.3532676} {Beyond opinion
  mining: Summarizing opinions of customer reviews}.
\newblock In \emph{{SIGIR} '22: The 45th International {ACM} {SIGIR} Conference
  on Research and Development in Information Retrieval, Madrid, Spain, July 11
  - 15, 2022}, pages 3447--3450. {ACM}.

\bibitem[{Andreas et~al.(2020)Andreas, Bufe, Burkett, Chen, Clausman, Crawford,
  Crim, DeLoach, Dorner, Eisner, Fang, Guo, Hall, Hayes, Hill, Ho, Iwaszuk,
  Jha, Klein, Krishnamurthy, Lanman, Liang, Lin, Lintsbakh, McGovern,
  Nisnevich, Pauls, Petters, Read, Roth, Roy, Rusak, Short, Slomin, Snyder,
  Striplin, Su, Tellman, Thomson, Vorobev, Witoszko, Wolfe, Wray, Zhang, and
  Zotov}]{data_SMCalFlow}
Jacob Andreas, John Bufe, David Burkett, Charles Chen, Josh Clausman, Jean
  Crawford, Kate Crim, Jordan DeLoach, Leah Dorner, Jason Eisner, Hao Fang,
  Alan Guo, David Hall, Kristin Hayes, Kellie Hill, Diana Ho, Wendy Iwaszuk,
  Smriti Jha, Dan Klein, Jayant Krishnamurthy, Theo Lanman, Percy Liang,
  Christopher~H. Lin, Ilya Lintsbakh, Andy McGovern, Aleksandr Nisnevich, Adam
  Pauls, Dmitrij Petters, Brent Read, Dan Roth, Subhro Roy, Jesse Rusak, Beth
  Short, Div Slomin, Ben Snyder, Stephon Striplin, Yu~Su, Zachary Tellman, Sam
  Thomson, Andrei Vorobev, Izabela Witoszko, Jason~Andrew Wolfe, Abby Wray,
  Yuchen Zhang, and Alexander Zotov. 2020.
\newblock \href {https://doi.org/10.1162/tacl\_a\_00333} {Task-oriented
  dialogue as dataflow synthesis}.
\newblock \emph{Trans. Assoc. Comput. Linguistics}, 8:556--571.

\bibitem[{Bar{-}Haim et~al.(2014)Bar{-}Haim, Dagan, and Szpektor}]{data_rte}
Roy Bar{-}Haim, Ido Dagan, and Idan Szpektor. 2014.
\newblock \href {https://doi.org/10.1007/978-3-642-45321-2\_19} {Benchmarking
  applied semantic inference: The {PASCAL} recognising textual entailment
  challenges}.
\newblock In \emph{Language, Culture, Computation. Computing - Theory and
  Technology - Essays Dedicated to Yaacov Choueka on the Occasion of His 75th
  Birthday, Part {I}}, volume 8001 of \emph{Lecture Notes in Computer Science},
  pages 409--424. Springer.

\bibitem[{Bender et~al.(2021)Bender, Gebru, McMillan{-}Major, and
  Shmitchell}]{language_model_bias}
Emily~M. Bender, Timnit Gebru, Angelina McMillan{-}Major, and Shmargaret
  Shmitchell. 2021.
\newblock \href {https://doi.org/10.1145/3442188.3445922} {On the dangers of
  stochastic parrots: Can language models be too big?}
\newblock In \emph{FAccT '21: 2021 {ACM} Conference on Fairness,
  Accountability, and Transparency, Virtual Event / Toronto, Canada, March
  3-10, 2021}, pages 610--623. {ACM}.

\bibitem[{Black et~al.(2021)Black, Gao, Wang, Leahy, and Biderman}]{gpt_neo}
Sid Black, Leo Gao, Phil Wang, Connor Leahy, and Stella Biderman. 2021.
\newblock \href {https://doi.org/10.5281/zenodo.5297715} {{GPT-Neo: Large Scale
  Autoregressive Language Modeling with Mesh-Tensorflow}}.

\bibitem[{Bowman et~al.(2015)Bowman, Angeli, Potts, and Manning}]{data_snli}
Samuel~R. Bowman, Gabor Angeli, Christopher Potts, and Christopher~D. Manning.
  2015.
\newblock \href {https://doi.org/10.18653/v1/d15-1075} {A large annotated
  corpus for learning natural language inference}.
\newblock In \emph{Proceedings of the 2015 Conference on Empirical Methods in
  Natural Language Processing, {EMNLP} 2015, Lisbon, Portugal, September 17-21,
  2015}, pages 632--642. The Association for Computational Linguistics.

\bibitem[{Brown et~al.(2020)Brown, Mann, Ryder, Subbiah, Kaplan, Dhariwal,
  Neelakantan, Shyam, Sastry, Askell, Agarwal, Herbert{-}Voss, Krueger,
  Henighan, Child, Ramesh, Ziegler, Wu, Winter, Hesse, Chen, Sigler, Litwin,
  Gray, Chess, Clark, Berner, McCandlish, Radford, Sutskever, and
  Amodei}]{gpt3}
Tom~B. Brown, Benjamin Mann, Nick Ryder, Melanie Subbiah, Jared Kaplan,
  Prafulla Dhariwal, Arvind Neelakantan, Pranav Shyam, Girish Sastry, Amanda
  Askell, Sandhini Agarwal, Ariel Herbert{-}Voss, Gretchen Krueger, Tom
  Henighan, Rewon Child, Aditya Ramesh, Daniel~M. Ziegler, Jeffrey Wu, Clemens
  Winter, Christopher Hesse, Mark Chen, Eric Sigler, Mateusz Litwin, Scott
  Gray, Benjamin Chess, Jack Clark, Christopher Berner, Sam McCandlish, Alec
  Radford, Ilya Sutskever, and Dario Amodei. 2020.
\newblock \href
  {https://proceedings.neurips.cc/paper/2020/hash/1457c0d6bfcb4967418bfb8ac142f64a-Abstract.html}
  {Language models are few-shot learners}.
\newblock In \emph{Advances in Neural Information Processing Systems 33: Annual
  Conference on Neural Information Processing Systems 2020, NeurIPS 2020,
  December 6-12, 2020, virtual}.

\bibitem[{Burges(2010)}]{lambdarank}
Christopher J.~C. Burges. 2010.
\newblock \href
  {http://research.microsoft.com/en-us/um/people/cburges/tech\_reports/MSR-TR-2010-82.pdf}
  {From {RankNet} to {LambdaRank} to {LambdaMART}: An overview}.
\newblock Technical report, Microsoft Research.

\bibitem[{Chen et~al.(2021)Chen, Tworek, Jun, Yuan, de~Oliveira~Pinto, Kaplan,
  Edwards, Burda, Joseph, Brockman, Ray, Puri, Krueger, Petrov, Khlaaf, Sastry,
  Mishkin, Chan, Gray, Ryder, Pavlov, Power, Kaiser, Bavarian, Winter, Tillet,
  Such, Cummings, Plappert, Chantzis, Barnes, Herbert{-}Voss, Guss, Nichol,
  Paino, Tezak, Tang, Babuschkin, Balaji, Jain, Saunders, Hesse, Carr, Leike,
  Achiam, Misra, Morikawa, Radford, Knight, Brundage, Murati, Mayer, Welinder,
  McGrew, Amodei, McCandlish, Sutskever, and Zaremba}]{codex}
Mark Chen, Jerry Tworek, Heewoo Jun, Qiming Yuan, Henrique~Ponde
  de~Oliveira~Pinto, Jared Kaplan, Harrison Edwards, Yuri Burda, Nicholas
  Joseph, Greg Brockman, Alex Ray, Raul Puri, Gretchen Krueger, Michael Petrov,
  Heidy Khlaaf, Girish Sastry, Pamela Mishkin, Brooke Chan, Scott Gray, Nick
  Ryder, Mikhail Pavlov, Alethea Power, Lukasz Kaiser, Mohammad Bavarian,
  Clemens Winter, Philippe Tillet, Felipe~Petroski Such, Dave Cummings,
  Matthias Plappert, Fotios Chantzis, Elizabeth Barnes, Ariel Herbert{-}Voss,
  William~Hebgen Guss, Alex Nichol, Alex Paino, Nikolas Tezak, Jie Tang, Igor
  Babuschkin, Suchir Balaji, Shantanu Jain, William Saunders, Christopher
  Hesse, Andrew~N. Carr, Jan Leike, Joshua Achiam, Vedant Misra, Evan Morikawa,
  Alec Radford, Matthew Knight, Miles Brundage, Mira Murati, Katie Mayer, Peter
  Welinder, Bob McGrew, Dario Amodei, Sam McCandlish, Ilya Sutskever, and
  Wojciech Zaremba. 2021.
\newblock \href {http://arxiv.org/abs/2107.03374} {Evaluating large language
  models trained on code}.
\newblock \emph{CoRR}, abs/2107.03374.

\bibitem[{Chen et~al.(2022)Chen, Zhao, Yu, McKeown, and
  He}]{sensitivity_and_accuracy}
Yanda Chen, Chen Zhao, Zhou Yu, Kathleen~R. McKeown, and He~He. 2022.
\newblock \href {https://doi.org/10.48550/arXiv.2209.07661} {On the relation
  between sensitivity and accuracy in in-context learning}.
\newblock \emph{CoRR}, abs/2209.07661.

\bibitem[{Cheng et~al.(2023)Cheng, Yang, Sun, Li, and
  Qiu}]{sentence_embedding_by_gpt3}
Qinyuan Cheng, Xiaogui Yang, Tianxiang Sun, Linyang Li, and Xipeng Qiu. 2023.
\newblock \href {https://doi.org/10.48550/arXiv.2305.01918} {Improving
  contrastive learning of sentence embeddings from {AI} feedback}.
\newblock \emph{CoRR}, abs/2305.01918.

\bibitem[{Cohan et~al.(2018)Cohan, Dernoncourt, Kim, Bui, Kim, Chang, and
  Goharian}]{data_pubmed}
Arman Cohan, Franck Dernoncourt, Doo~Soon Kim, Trung Bui, Seokhwan Kim, Walter
  Chang, and Nazli Goharian. 2018.
\newblock \href {https://doi.org/10.18653/v1/N18-2097} {A discourse-aware
  attention model for abstractive summarization of long documents}.
\newblock In \emph{Proceedings of the 2018 Conference of the North {A}merican
  Chapter of the Association for Computational Linguistics: Human Language
  Technologies, Volume 2 (Short Papers)}, pages 615--621, New Orleans,
  Louisiana. Association for Computational Linguistics.

\bibitem[{Conneau and Lample(2019)}]{DBLP:conf/nips/ConneauL19}
Alexis Conneau and Guillaume Lample. 2019.
\newblock \href
  {https://proceedings.neurips.cc/paper/2019/hash/c04c19c2c2474dbf5f7ac4372c5b9af1-Abstract.html}
  {Cross-lingual language model pretraining}.
\newblock In \emph{Advances in Neural Information Processing Systems 32: Annual
  Conference on Neural Information Processing Systems 2019, NeurIPS 2019,
  December 8-14, 2019, Vancouver, BC, Canada}, pages 7057--7067.

\bibitem[{Das et~al.(2021)Das, Zaheer, Thai, Godbole, Perez, Lee, Tan,
  Polymenakos, and McCallum}]{cbr}
Rajarshi Das, Manzil Zaheer, Dung Thai, Ameya Godbole, Ethan Perez, Jay~Yoon
  Lee, Lizhen Tan, Lazaros Polymenakos, and Andrew McCallum. 2021.
\newblock \href {https://doi.org/10.18653/v1/2021.emnlp-main.755} {Case-based
  reasoning for natural language queries over knowledge bases}.
\newblock In \emph{Proceedings of the 2021 Conference on Empirical Methods in
  Natural Language Processing, {EMNLP} 2021, Virtual Event / Punta Cana,
  Dominican Republic, 7-11 November, 2021}, pages 9594--9611. Association for
  Computational Linguistics.

\bibitem[{Devlin et~al.(2019)Devlin, Chang, Lee, and Toutanova}]{bert}
Jacob Devlin, Ming{-}Wei Chang, Kenton Lee, and Kristina Toutanova. 2019.
\newblock \href {https://doi.org/10.18653/v1/n19-1423} {{BERT:} pre-training of
  deep bidirectional transformers for language understanding}.
\newblock In \emph{Proceedings of the 2019 Conference of the North American
  Chapter of the Association for Computational Linguistics: Human Language
  Technologies, {NAACL-HLT} 2019, Minneapolis, MN, USA, June 2-7, 2019, Volume
  1 (Long and Short Papers)}, pages 4171--4186. Association for Computational
  Linguistics.

\bibitem[{Dong et~al.(2022)Dong, Li, Dai, Zheng, Wu, Chang, Sun, Xu, Li, and
  Sui}]{icl_survey}
Qingxiu Dong, Lei Li, Damai Dai, Ce~Zheng, Zhiyong Wu, Baobao Chang, Xu~Sun,
  Jingjing Xu, Lei Li, and Zhifang Sui. 2022.
\newblock \href {http://arxiv.org/abs/2301.00234} {A survey for in-context
  learning}.

\bibitem[{Dušek et~al.(2019)Dušek, Howcroft, and Rieser}]{data_e2e}
Ondřej Dušek, David~M Howcroft, and Verena Rieser. 2019.
\newblock \href {https://www.aclweb.org/anthology/W19-8652/} {Semantic {Noise}
  {Matters} for {Neural} {Natural} {Language} {Generation}}.
\newblock In \emph{Proceedings of the 12th {International} {Conference} on
  {Natural} {Language} {Generation} ({INLG} 2019)}, pages 421--426, Tokyo,
  Japan.

\bibitem[{He et~al.(2021)He, Liu, Gao, and Chen}]{deberta}
Pengcheng He, Xiaodong Liu, Jianfeng Gao, and Weizhu Chen. 2021.
\newblock \href {https://openreview.net/forum?id=XPZIaotutsD} {Deberta:
  decoding-enhanced bert with disentangled attention}.
\newblock In \emph{9th International Conference on Learning Representations,
  {ICLR} 2021, Virtual Event, Austria, May 3-7, 2021}. OpenReview.net.

\bibitem[{Hermann et~al.(2015)Hermann, Kocisk{\'{y}}, Grefenstette, Espeholt,
  Kay, Suleyman, and Blunsom}]{data_cnn_dm}
Karl~Moritz Hermann, Tom{\'{a}}s Kocisk{\'{y}}, Edward Grefenstette, Lasse
  Espeholt, Will Kay, Mustafa Suleyman, and Phil Blunsom. 2015.
\newblock \href
  {https://proceedings.neurips.cc/paper/2015/hash/afdec7005cc9f14302cd0474fd0f3c96-Abstract.html}
  {Teaching machines to read and comprehend}.
\newblock In \emph{Advances in Neural Information Processing Systems 28: Annual
  Conference on Neural Information Processing Systems 2015, December 7-12,
  2015, Montreal, Quebec, Canada}, pages 1693--1701.

\bibitem[{Hu et~al.(2022)Hu, Lee, Xie, Yu, Smith, and Ostendorf}]{ic_dst}
Yushi Hu, Chia{-}Hsuan Lee, Tianbao Xie, Tao Yu, Noah~A. Smith, and Mari
  Ostendorf. 2022.
\newblock \href {https://doi.org/10.48550/arXiv.2203.08568} {In-context
  learning for few-shot dialogue state tracking}.
\newblock \emph{CoRR}, abs/2203.08568.

\bibitem[{Huang et~al.(2019)Huang, Le~Bras, Bhagavatula, and
  Choi}]{data_cosmos}
Lifu Huang, Ronan Le~Bras, Chandra Bhagavatula, and Yejin Choi. 2019.
\newblock \href {https://doi.org/10.18653/v1/D19-1243} {Cosmos {QA}: Machine
  reading comprehension with contextual commonsense reasoning}.
\newblock In \emph{Proceedings of the 2019 Conference on Empirical Methods in
  Natural Language Processing and the 9th International Joint Conference on
  Natural Language Processing (EMNLP-IJCNLP)}, pages 2391--2401, Hong Kong,
  China. Association for Computational Linguistics.

\bibitem[{Jiang et~al.(2020)Jiang, Maddela, Lan, Zhong, and
  Xu}]{data_wiki_auto}
Chao Jiang, Mounica Maddela, Wuwei Lan, Yang Zhong, and Wei Xu. 2020.
\newblock \href {https://doi.org/10.18653/v1/2020.acl-main.709} {Neural {CRF}
  model for sentence alignment in text simplification}.
\newblock In \emph{Proceedings of the 58th Annual Meeting of the Association
  for Computational Linguistics}, pages 7943--7960, Online. Association for
  Computational Linguistics.

\bibitem[{Johnson et~al.(2021)Johnson, Douze, and J{\'{e}}gou}]{faiss}
Jeff Johnson, Matthijs Douze, and Herv{\'{e}} J{\'{e}}gou. 2021.
\newblock \href {https://doi.org/10.1109/TBDATA.2019.2921572} {Billion-scale
  similarity search with gpus}.
\newblock \emph{{IEEE} Trans. Big Data}, 7(3):535--547.

\bibitem[{Karpukhin et~al.(2020)Karpukhin, Oguz, Min, Lewis, Wu, Edunov, Chen,
  and Yih}]{dpr}
Vladimir Karpukhin, Barlas Oguz, Sewon Min, Patrick S.~H. Lewis, Ledell Wu,
  Sergey Edunov, Danqi Chen, and Wen{-}tau Yih. 2020.
\newblock \href {https://doi.org/10.18653/v1/2020.emnlp-main.550} {Dense
  passage retrieval for open-domain question answering}.
\newblock In \emph{Proceedings of the 2020 Conference on Empirical Methods in
  Natural Language Processing, {EMNLP} 2020, Online, November 16-20, 2020},
  pages 6769--6781. Association for Computational Linguistics.

\bibitem[{Kim et~al.(2019)Kim, Kim, and Kim}]{data_reddit}
Byeongchang Kim, Hyunwoo Kim, and Gunhee Kim. 2019.
\newblock \href {https://doi.org/10.18653/v1/N19-1260} {Abstractive
  summarization of {R}eddit posts with multi-level memory networks}.
\newblock In \emph{Proceedings of the 2019 Conference of the North {A}merican
  Chapter of the Association for Computational Linguistics: Human Language
  Technologies, Volume 1 (Long and Short Papers)}, pages 2519--2531,
  Minneapolis, Minnesota. Association for Computational Linguistics.

\bibitem[{Lehmann et~al.(2015)Lehmann, Isele, Jakob, Jentzsch, Kontokostas,
  Mendes, Hellmann, Morsey, van Kleef, Auer, and Bizer}]{data_dbpedia}
Jens Lehmann, Robert Isele, Max Jakob, Anja Jentzsch, Dimitris Kontokostas,
  Pablo~N. Mendes, Sebastian Hellmann, Mohamed Morsey, Patrick van Kleef,
  S{\"{o}}ren Auer, and Christian Bizer. 2015.
\newblock \href {https://doi.org/10.3233/SW-140134} {Dbpedia - {A} large-scale,
  multilingual knowledge base extracted from wikipedia}.
\newblock \emph{Semantic Web}, 6(2):167--195.

\bibitem[{Li et~al.(2021)Li, Arora, Chen, Gupta, Gupta, and Mehdad}]{data_mtop}
Haoran Li, Abhinav Arora, Shuohui Chen, Anchit Gupta, Sonal Gupta, and Yashar
  Mehdad. 2021.
\newblock \href {https://doi.org/10.18653/v1/2021.eacl-main.257} {{MTOP}: A
  comprehensive multilingual task-oriented semantic parsing benchmark}.
\newblock In \emph{Proceedings of the 16th Conference of the European Chapter
  of the Association for Computational Linguistics: Main Volume}, pages
  2950--2962, Online. Association for Computational Linguistics.

\bibitem[{Li and Qiu(2023{\natexlab{a}})}]{supporting_examples}
Xiaonan Li and Xipeng Qiu. 2023{\natexlab{a}}.
\newblock \href {https://doi.org/10.48550/arXiv.2302.13539} {Finding supporting
  examples for in-context learning}.
\newblock \emph{CoRR}, abs/2302.13539.

\bibitem[{Li and Qiu(2023{\natexlab{b}})}]{mot}
Xiaonan Li and Xipeng Qiu. 2023{\natexlab{b}}.
\newblock \href {http://arxiv.org/abs/2305.05181} {Mot: Pre-thinking and
  recalling enable chatgpt to self-improve with memory-of-thoughts}.

\bibitem[{Lin et~al.(2020)Lin, Zhou, Shen, Zhou, Bhagavatula, Choi, and
  Ren}]{data_common_gen}
Bill~Yuchen Lin, Wangchunshu Zhou, Ming Shen, Pei Zhou, Chandra Bhagavatula,
  Yejin Choi, and Xiang Ren. 2020.
\newblock \href {https://doi.org/10.18653/v1/2020.findings-emnlp.165}
  {{C}ommon{G}en: A constrained text generation challenge for generative
  commonsense reasoning}.
\newblock In \emph{Findings of the Association for Computational Linguistics:
  EMNLP 2020}, pages 1823--1840, Online. Association for Computational
  Linguistics.

\bibitem[{Liu et~al.(2022)Liu, Shen, Zhang, Dolan, Carin, and
  Chen}]{what_is_good_example_for_gpt3}
Jiachang Liu, Dinghan Shen, Yizhe Zhang, Bill Dolan, Lawrence Carin, and Weizhu
  Chen. 2022.
\newblock \href {https://doi.org/10.18653/v1/2022.deelio-1.10} {What makes good
  in-context examples for gpt-3?}
\newblock In \emph{Proceedings of Deep Learning Inside Out: The 3rd Workshop on
  Knowledge Extraction and Integration for Deep Learning Architectures,
  DeeLIO@ACL 2022, Dublin, Ireland and Online, May 27, 2022}, pages 100--114.
  Association for Computational Linguistics.

\bibitem[{Liu et~al.(2019)Liu, Ott, Goyal, Du, Joshi, Chen, Levy, Lewis,
  Zettlemoyer, and Stoyanov}]{roberta}
Yinhan Liu, Myle Ott, Naman Goyal, Jingfei Du, Mandar Joshi, Danqi Chen, Omer
  Levy, Mike Lewis, Luke Zettlemoyer, and Veselin Stoyanov. 2019.
\newblock \href {http://arxiv.org/abs/1907.11692} {Roberta: {A} robustly
  optimized {BERT} pretraining approach}.
\newblock \emph{CoRR}, abs/1907.11692.

\bibitem[{Loshchilov and Hutter(2019)}]{adamw}
Ilya Loshchilov and Frank Hutter. 2019.
\newblock \href {https://openreview.net/forum?id=Bkg6RiCqY7} {Decoupled weight
  decay regularization}.
\newblock In \emph{7th International Conference on Learning Representations,
  {ICLR} 2019, New Orleans, LA, USA, May 6-9, 2019}. OpenReview.net.

\bibitem[{Lu et~al.(2021)Lu, Guo, Ren, Huang, Svyatkovskiy, Blanco, Clement,
  Drain, Jiang, Tang, Li, Zhou, Shou, Zhou, Tufano, Gong, Zhou, Duan,
  Sundaresan, Deng, Fu, and Liu}]{data_codexglue}
Shuai Lu, Daya Guo, Shuo Ren, Junjie Huang, Alexey Svyatkovskiy, Ambrosio
  Blanco, Colin~B. Clement, Dawn Drain, Daxin Jiang, Duyu Tang, Ge~Li, Lidong
  Zhou, Linjun Shou, Long Zhou, Michele Tufano, Ming Gong, Ming Zhou, Nan Duan,
  Neel Sundaresan, Shao~Kun Deng, Shengyu Fu, and Shujie Liu. 2021.
\newblock \href
  {https://datasets-benchmarks-proceedings.neurips.cc/paper/2021/hash/c16a5320fa475530d9583c34fd356ef5-Abstract-round1.html}
  {Codexglue: {A} machine learning benchmark dataset for code understanding and
  generation}.
\newblock In \emph{Proceedings of the Neural Information Processing Systems
  Track on Datasets and Benchmarks 1, NeurIPS Datasets and Benchmarks 2021,
  December 2021, virtual}.

\bibitem[{Lu et~al.(2022)Lu, Bartolo, Moore, Riedel, and
  Stenetorp}]{icl_ordering}
Yao Lu, Max Bartolo, Alastair Moore, Sebastian Riedel, and Pontus Stenetorp.
  2022.
\newblock \href {https://doi.org/10.18653/v1/2022.acl-long.556} {Fantastically
  ordered prompts and where to find them: Overcoming few-shot prompt order
  sensitivity}.
\newblock In \emph{Proceedings of the 60th Annual Meeting of the Association
  for Computational Linguistics (Volume 1: Long Papers)}, pages 8086--8098,
  Dublin, Ireland. Association for Computational Linguistics.

\bibitem[{McAuley and Leskovec(2013)}]{data_amazon}
Julian~J. McAuley and Jure Leskovec. 2013.
\newblock \href {https://doi.org/10.1145/2507157.2507163} {Hidden factors and
  hidden topics: understanding rating dimensions with review text}.
\newblock In \emph{Seventh {ACM} Conference on Recommender Systems, RecSys '13,
  Hong Kong, China, October 12-16, 2013}, pages 165--172. {ACM}.

\bibitem[{Min et~al.(2022)Min, Lewis, Hajishirzi, and Zettlemoyer}]{channel}
Sewon Min, Mike Lewis, Hannaneh Hajishirzi, and Luke Zettlemoyer. 2022.
\newblock \href {https://doi.org/10.18653/v1/2022.acl-long.365} {Noisy channel
  language model prompting for few-shot text classification}.
\newblock In \emph{Proceedings of the 60th Annual Meeting of the Association
  for Computational Linguistics (Volume 1: Long Papers), {ACL} 2022, Dublin,
  Ireland, May 22-27, 2022}, pages 5316--5330. Association for Computational
  Linguistics.

\bibitem[{Mostafazadeh et~al.(2016)Mostafazadeh, Chambers, He, Parikh, Batra,
  Vanderwende, Kohli, and Allen}]{data_roc}
Nasrin Mostafazadeh, Nathanael Chambers, Xiaodong He, Devi Parikh, Dhruv Batra,
  Lucy Vanderwende, Pushmeet Kohli, and James Allen. 2016.
\newblock \href {https://doi.org/10.18653/v1/N16-1098} {A corpus and cloze
  evaluation for deeper understanding of commonsense stories}.
\newblock In \emph{Proceedings of the 2016 Conference of the North {A}merican
  Chapter of the Association for Computational Linguistics: Human Language
  Technologies}, pages 839--849, San Diego, California. Association for
  Computational Linguistics.

\bibitem[{Naji(2012)}]{data_twitter_sentiment}
Ibrahim Naji. 2012.
\newblock {TSATC: Twitter Sentiment Analysis Training Corpus}.
\newblock In \emph{thinknook}.

\bibitem[{Nan et~al.(2021)Nan, Radev, Zhang, Rau, Sivaprasad, Hsieh, Tang,
  Vyas, Verma, Krishna, Liu, Irwanto, Pan, Rahman, Zaidi, Mutuma, Tarabar,
  Gupta, Yu, Tan, Lin, Xiong, Socher, and Rajani}]{data_dart}
Linyong Nan, Dragomir Radev, Rui Zhang, Amrit Rau, Abhinand Sivaprasad,
  Chiachun Hsieh, Xiangru Tang, Aadit Vyas, Neha Verma, Pranav Krishna,
  Yangxiaokang Liu, Nadia Irwanto, Jessica Pan, Faiaz Rahman, Ahmad Zaidi,
  Mutethia Mutuma, Yasin Tarabar, Ankit Gupta, Tao Yu, Yi~Chern Tan,
  Xi~Victoria Lin, Caiming Xiong, Richard Socher, and Nazneen~Fatema Rajani.
  2021.
\newblock \href {https://doi.org/10.18653/v1/2021.naacl-main.37} {{DART}:
  Open-domain structured data record to text generation}.
\newblock In \emph{Proceedings of the 2021 Conference of the North American
  Chapter of the Association for Computational Linguistics: Human Language
  Technologies}, pages 432--447, Online. Association for Computational
  Linguistics.

\bibitem[{Ouyang et~al.(2022)Ouyang, Wu, Jiang, Almeida, Wainwright, Mishkin,
  Zhang, Agarwal, Slama, Ray, Schulman, Hilton, Kelton, Miller, Simens, Askell,
  Welinder, Christiano, Leike, and Lowe}]{instruct_gpt}
Long Ouyang, Jeff Wu, Xu~Jiang, Diogo Almeida, Carroll~L. Wainwright, Pamela
  Mishkin, Chong Zhang, Sandhini Agarwal, Katarina Slama, Alex Ray, John
  Schulman, Jacob Hilton, Fraser Kelton, Luke Miller, Maddie Simens, Amanda
  Askell, Peter Welinder, Paul~F. Christiano, Jan Leike, and Ryan Lowe. 2022.
\newblock \href {https://doi.org/10.48550/arXiv.2203.02155} {Training language
  models to follow instructions with human feedback}.
\newblock \emph{CoRR}, abs/2203.02155.

\bibitem[{Pang and Lee(2004)}]{data_subj}
Bo~Pang and Lillian Lee. 2004.
\newblock \href {https://doi.org/10.3115/1218955.1218990} {A sentimental
  education: Sentiment analysis using subjectivity summarization based on
  minimum cuts}.
\newblock In \emph{Proceedings of the 42nd Annual Meeting of the Association
  for Computational Linguistics, 21-26 July, 2004, Barcelona, Spain}, pages
  271--278. {ACL}.

\bibitem[{Pang and Lee(2005)}]{data_mr}
Bo~Pang and Lillian Lee. 2005.
\newblock \href {https://doi.org/10.3115/1219840.1219855} {Seeing stars:
  Exploiting class relationships for sentiment categorization with respect to
  rating scales}.
\newblock In \emph{{ACL} 2005, 43rd Annual Meeting of the Association for
  Computational Linguistics, Proceedings of the Conference, 25-30 June 2005,
  University of Michigan, {USA}}, pages 115--124. The Association for Computer
  Linguistics.

\bibitem[{Poesia et~al.(2022)Poesia, Polozov, Le, Tiwari, Soares, Meek, and
  Gulwani}]{reliable_code_generation}
Gabriel Poesia, Alex Polozov, Vu~Le, Ashish Tiwari, Gustavo Soares, Christopher
  Meek, and Sumit Gulwani. 2022.
\newblock \href {https://openreview.net/forum?id=KmtVD97J43e} {Synchromesh:
  Reliable code generation from pre-trained language models}.
\newblock In \emph{The Tenth International Conference on Learning
  Representations, {ICLR} 2022, Virtual Event, April 25-29, 2022}.
  OpenReview.net.

\bibitem[{Radford et~al.(2018)Radford, Wu, Child, Luan, Amodei, and
  Sutskever}]{gpt2}
Alec Radford, Jeffrey Wu, Rewon Child, David Luan, Dario Amodei, and Ilya
  Sutskever. 2018.
\newblock \href
  {https://d4mucfpksywv.cloudfront.net/better-language-models/language-models.pdf}
  {Language models are unsupervised multitask learners}.

\bibitem[{Reimers and Gurevych(2019{\natexlab{a}})}]{sentence_bert}
Nils Reimers and Iryna Gurevych. 2019{\natexlab{a}}.
\newblock \href {https://doi.org/10.18653/v1/D19-1410} {Sentence-{BERT}:
  Sentence embeddings using {S}iamese {BERT}-networks}.
\newblock In \emph{Proceedings of the 2019 Conference on Empirical Methods in
  Natural Language Processing and the 9th International Joint Conference on
  Natural Language Processing (EMNLP-IJCNLP)}, pages 3982--3992, Hong Kong,
  China. Association for Computational Linguistics.

\bibitem[{Reimers and Gurevych(2019{\natexlab{b}})}]{sentencebert}
Nils Reimers and Iryna Gurevych. 2019{\natexlab{b}}.
\newblock \href {https://doi.org/10.18653/v1/D19-1410} {Sentence-bert: Sentence
  embeddings using siamese bert-networks}.
\newblock In \emph{Proceedings of the 2019 Conference on Empirical Methods in
  Natural Language Processing and the 9th International Joint Conference on
  Natural Language Processing, {EMNLP-IJCNLP} 2019, Hong Kong, China, November
  3-7, 2019}, pages 3980--3990. Association for Computational Linguistics.

\bibitem[{Robertson and Zaragoza(2009)}]{bm25}
Stephen~E. Robertson and Hugo Zaragoza. 2009.
\newblock \href {https://doi.org/10.1561/1500000019} {The probabilistic
  relevance framework: {BM25} and beyond}.
\newblock \emph{Found. Trends Inf. Retr.}, 3(4):333--389.

\bibitem[{Roemmele et~al.(2011)Roemmele, Bejan, and Gordon}]{data_copa}
Melissa Roemmele, Cosmin~Adrian Bejan, and Andrew~S Gordon. 2011.
\newblock \href
  {https://people.ict.usc.edu/~gordon/publications/AAAI-SPRING11A.PDF} {Choice
  of plausible alternatives: An evaluation of commonsense causal reasoning}.
\newblock In \emph{2011 AAAI Spring Symposium Series}.

\bibitem[{Rubin et~al.(2022)Rubin, Herzig, and Berant}]{epr}
Ohad Rubin, Jonathan Herzig, and Jonathan Berant. 2022.
\newblock \href {https://doi.org/10.18653/v1/2022.naacl-main.191} {Learning to
  retrieve prompts for in-context learning}.
\newblock In \emph{Proceedings of the 2022 Conference of the North American
  Chapter of the Association for Computational Linguistics: Human Language
  Technologies, {NAACL} 2022, Seattle, WA, United States, July 10-15, 2022},
  pages 2655--2671. Association for Computational Linguistics.

\bibitem[{Shi et~al.(2022)Shi, Zhang, Bai, and
  Lin}]{cross_lingual_icl_for_text_sql}
Peng Shi, Rui Zhang, He~Bai, and Jimmy Lin. 2022.
\newblock \href {https://doi.org/10.48550/arXiv.2210.13693} {{XRICL:}
  cross-lingual retrieval-augmented in-context learning for cross-lingual
  text-to-sql semantic parsing}.
\newblock \emph{CoRR}, abs/2210.13693.

\bibitem[{Socher et~al.(2013)Socher, Perelygin, Wu, Chuang, Manning, Ng, and
  Potts}]{data_sst2_and_sst5}
Richard Socher, Alex Perelygin, Jean Wu, Jason Chuang, Christopher~D. Manning,
  Andrew~Y. Ng, and Christopher Potts. 2013.
\newblock \href {https://aclanthology.org/D13-1170/} {Recursive deep models for
  semantic compositionality over a sentiment treebank}.
\newblock In \emph{Proceedings of the 2013 Conference on Empirical Methods in
  Natural Language Processing, {EMNLP} 2013, 18-21 October 2013, Grand Hyatt
  Seattle, Seattle, Washington, USA, {A} meeting of SIGDAT, a Special Interest
  Group of the {ACL}}, pages 1631--1642. {ACL}.

\bibitem[{Su et~al.(2022)Su, Shi, Kasai, Yizhong~Wang, Ostendorf, tau Yih,
  Smith, Zettlemoyer, and Yu}]{instructor}
Hongjin Su, Weijia Shi, Jungo Kasai, Yushi~Hu Yizhong~Wang, Mari Ostendorf, Wen
  tau Yih, Noah~A. Smith, Luke Zettlemoyer, and Tao Yu. 2022.
\newblock \href {https://arxiv.org/abs/2212.09741} {One embedder, any task:
  Instruction-finetuned text embeddings}.

\bibitem[{Vaswani et~al.(2017)Vaswani, Shazeer, Parmar, Uszkoreit, Jones,
  Gomez, Kaiser, and Polosukhin}]{transformer}
Ashish Vaswani, Noam Shazeer, Niki Parmar, Jakob Uszkoreit, Llion Jones,
  Aidan~N Gomez, \L~ukasz Kaiser, and Illia Polosukhin. 2017.
\newblock \href
  {https://proceedings.neurips.cc/paper/2017/file/3f5ee243547dee91fbd053c1c4a845aa-Paper.pdf}
  {Attention is all you need}.
\newblock In \emph{Advances in Neural Information Processing Systems},
  volume~30. Curran Associates, Inc.

\bibitem[{Voorhees and Tice(2000)}]{data_trec}
Ellen~M. Voorhees and Dawn~M. Tice. 2000.
\newblock \href {https://doi.org/10.1145/345508.345577} {Building a question
  answering test collection}.
\newblock In \emph{{SIGIR} 2000: Proceedings of the 23rd Annual International
  {ACM} {SIGIR} Conference on Research and Development in Information
  Retrieval, July 24-28, 2000, Athens, Greece}, pages 200--207. {ACM}.

\bibitem[{Wang et~al.(2019{\natexlab{a}})Wang, Singh, Michael, Hill, Levy, and
  Bowman}]{data_glue}
Alex Wang, Amanpreet Singh, Julian Michael, Felix Hill, Omer Levy, and
  Samuel~R. Bowman. 2019{\natexlab{a}}.
\newblock \href {https://openreview.net/forum?id=rJ4km2R5t7} {{GLUE:} {A}
  multi-task benchmark and analysis platform for natural language
  understanding}.
\newblock In \emph{7th International Conference on Learning Representations,
  {ICLR} 2019, New Orleans, LA, USA, May 6-9, 2019}. OpenReview.net.

\bibitem[{Wang and Komatsuzaki(2021)}]{gpt_j}
Ben Wang and Aran Komatsuzaki. 2021.
\newblock {GPT-J-6B: A 6 Billion Parameter Autoregressive Language Model}.
\newblock \url{https://github.com/kingoflolz/mesh-transformer-jax}.

\bibitem[{Wang et~al.(2019{\natexlab{b}})Wang, Liang, Zhang, Li, and
  Gao}]{data_come_comv}
Cunxiang Wang, Shuailong Liang, Yue Zhang, Xiaonan Li, and Tian Gao.
  2019{\natexlab{b}}.
\newblock \href {https://doi.org/10.18653/v1/p19-1393} {Does it make sense? and
  why? {A} pilot study for sense making and explanation}.
\newblock In \emph{Proceedings of the 57th Conference of the Association for
  Computational Linguistics, {ACL} 2019, Florence, Italy, July 28- August 2,
  2019, Volume 1: Long Papers}, pages 4020--4026. Association for Computational
  Linguistics.

\bibitem[{Williams et~al.(2018)Williams, Nangia, and Bowman}]{data_mnli}
Adina Williams, Nikita Nangia, and Samuel~R. Bowman. 2018.
\newblock \href {https://doi.org/10.18653/v1/n18-1101} {A broad-coverage
  challenge corpus for sentence understanding through inference}.
\newblock In \emph{Proceedings of the 2018 Conference of the North American
  Chapter of the Association for Computational Linguistics: Human Language
  Technologies, {NAACL-HLT} 2018, New Orleans, Louisiana, USA, June 1-6, 2018,
  Volume 1 (Long Papers)}, pages 1112--1122. Association for Computational
  Linguistics.

\bibitem[{Wolfson et~al.(2020)Wolfson, Geva, Gupta, Goldberg, Gardner, Deutch,
  and Berant}]{data_break}
Tomer Wolfson, Mor Geva, Ankit Gupta, Yoav Goldberg, Matt Gardner, Daniel
  Deutch, and Jonathan Berant. 2020.
\newblock \href {https://doi.org/10.1162/tacl\_a\_00309} {Break it down: {A}
  question understanding benchmark}.
\newblock \emph{Trans. Assoc. Comput. Linguistics}, 8:183--198.

\bibitem[{Xiong et~al.(2021)Xiong, Xiong, Li, Tang, Liu, Bennett, Ahmed, and
  Overwijk}]{ance}
Lee Xiong, Chenyan Xiong, Ye~Li, Kwok{-}Fung Tang, Jialin Liu, Paul~N. Bennett,
  Junaid Ahmed, and Arnold Overwijk. 2021.
\newblock \href {https://openreview.net/forum?id=zeFrfgyZln} {Approximate
  nearest neighbor negative contrastive learning for dense text retrieval}.
\newblock In \emph{9th International Conference on Learning Representations,
  {ICLR} 2021, Virtual Event, Austria, May 3-7, 2021}. OpenReview.net.

\bibitem[{Xu et~al.(2023)Xu, Wang, Mao, Lyu, She, and Zhang}]{knn_prompting}
Benfeng Xu, Quan Wang, Zhendong Mao, Yajuan Lyu, Qiaoqiao She, and Yongdong
  Zhang. 2023.
\newblock \href {http://arxiv.org/abs/2303.13824} {$k$nn prompting:
  Beyond-context learning with calibration-free nearest neighbor inference}.

\bibitem[{Zhang et~al.(2015)Zhang, Zhao, and LeCun}]{data_yelp_agnews_yahoo}
Xiang Zhang, Junbo~Jake Zhao, and Yann LeCun. 2015.
\newblock \href
  {https://proceedings.neurips.cc/paper/2015/hash/250cf8b51c773f3f8dc8b4be867a9a02-Abstract.html}
  {Character-level convolutional networks for text classification}.
\newblock In \emph{Advances in Neural Information Processing Systems 28: Annual
  Conference on Neural Information Processing Systems 2015, December 7-12,
  2015, Montreal, Quebec, Canada}, pages 649--657.

\bibitem[{Zhang et~al.(2022)Zhang, Feng, and Tan}]{active_select_for_icl}
Yiming Zhang, Shi Feng, and Chenhao Tan. 2022.
\newblock \href {https://doi.org/10.48550/arXiv.2211.04486} {Active example
  selection for in-context learning}.
\newblock \emph{CoRR}, abs/2211.04486.

\end{thebibliography}
\bibliographystyle{acl_natbib}

\clearpage
\appendix
\section{Task Overview}
\label{appendix_task_overview}
We show each task's 1. input/output domain 2. statistics and evaluation metric 3. instruction, inference template and example cases in Table~\ref{tab_task_intro},~\ref{tab_statistics} and \ref{tab_templates}, respectively. For the dataset which has publicly available test data, we use the test data for evaluation, like SST-2, SST-5, MTOP, etc. For the others like BREAK and SMCalFlow, we follow previous work~\citep{epr} and use the dev data for evaluation. 
For training efficiency, we manually limit the training examples of UDR. Specifically, for the classification task whose training set size is > 30000, we randomly sample a 30000 subset for UDR's training. For the generation task whose training set size is > 100000, we randomly sample a 100000 subset for UDR's training. In the pilot experiment, we find such a strategy will not cause significant performance degradation. At the inference stage, we use the full training set as demonstrations' pool. Restricted by computational resources, we randomly sample a test set of 3000 samples for evaluation on these tasks: Amazon, Yelp, AGNews DBPedia and Yahoo.

\begin{table}[]
\small
\centering
\begin{tabular}{@{}lccc@{}}
\toprule
Template          & MR   & Yahoo                       & Subj \\ \midrule
Original Template & 85.2 & 67.5                        & 95.0   \\
Template 1        & 85.1 & 67.8                        & 94.8 \\
Template 2        & 85.7 & {\color[HTML]{333333} 67.1} & 94.8 \\
Template 3        & 85.4 & 67.2                        & 95.2 \\ \bottomrule
\end{tabular}
\caption{UDR's perfromance under different inference templates. For MR, the original template and template 1, 2, 3 are ``It was [Verbalizer]'', ``A [Verbalizer] One'', ``All in all [Verbalizer] .'', ``A [Verbalizer] one .'', respectively. The verbalizers are [``great'', ``terrible'']. For Yahoo, the original template and template 1, 2, 3 are ``Topic: [Verbalizer]'', ``Subject: [Verbalizer]'', ``This is about [Verbalizer] .'', ``It is about [Verbalizer] .'', respectively. The verbalizers are [``Society \& Culture'', ``Science \& Mathematics, $\cdots$]. For Subj, the original template and template 1, 2, 3 are ``It's [verbalizer] .'', ``This is [Verbalizer]'', ``It's all [Verbalizer].'', ``Is it [Verbalizer] ?'', respectively. The Verbalizers are [``subjective'', ``objective'']. These templates are from previous works~\citep{channel} and for more detals please refer to Table~\ref{tab_templates}.
}
\label{tab_different_tempalte}

\end{table}

\begin{table}[]
\small
\centering
\begin{tabular}{@{}lc@{}}
\toprule
\multicolumn{2}{c}{Hyper-parameters} \\ \midrule
Optimizer                    & AdamW \\
Warmup Steps                 & 500   \\
Learning Rate                & 1e-4  \\
Batch Size                   & 128   \\
Loss Weight                  & 0.8   \\
Iteration Number                & 3     \\
Scoring Candidates Num ($K$)   & 50    \\
Training Candidates Num ($l$)  & 8     \\ \bottomrule
\end{tabular}
\caption{Hyper-parameters.}
\label{tab_hypaer_parameters}
\end{table}

\section{Implementation Details and Hyper-Parameters}
\label{appendix_implementation_details_and_hyper_parameters}
We follow ~\citet{epr} to use GPT-Neo-2.7B~\citep{gpt_neo} as the scoring LM and the inference LM for most experiments in the paper unless otherwise specified. Following EPR~\citep{epr} and DPR \citep{dpr}, we initialize $E_q$ and $E_d$ as two separate ``BERT base uncased'' encoders~\citep{bert}. Thus the total number of parameters of UDR is about 220M. We use 8 NVIDIA A100s-80GB
to train UDR for up to 30 epochs before iteratively mining candidates. And then we train UDR for 10 epochs at each iteration. The whole training pipeline including scoring candidates takes about 8 days. In the pilot experiment, we select the number of training epochs through the average performance on validation set on single-task SST-2, TREC, MTOP, Java code summarization, WikiAuto and DART. We set the number of iterations as 3. We follow EPR~\citep{epr} to set learning rate and batch size as 1e-4 and 128 and we use AdamW~\citep{adamw} as the optimizer. We list the overall hyper-parameters in Table~\ref{tab_hypaer_parameters}. On each task, we use one specific template for scoring and inference (see Table~\ref{tab_templates}). For fair comparison, we train DR-Target, EPR and UDR under the same hyper-parameter and report their average performance under three random seeds. 
\paragraph{The initialization of UDR's candidates}
For classification and multi-choice tasks, we initialize candidates as those examples that have similar input with $x$ by BM25. For generation tasks, similarly, we initialize candidates as those of similar targets with $x$'s, inspired by previous work~\citep{epr}.

\paragraph{The Quantity of EPR's Candidates}
Since UDR's training needs to score iteratively mined candidates and thus has to score more candidates than EPR, we also run experiments on EPR with the same candidate quantities of UDR. But we find increasing the candidates of EPR instead slightly hurts its overall performance, which is consistent with its original paper~\citep{epr}. Thus for EPR, we use the same number of candidates as its original paper.

\section{Performance across varying inference templates}
\label{appendix_different_inference_template}
For UDR, we use one specific template when scoring candidates and here we evaluate UDR's transferability across different inference templates on MR, Yahoo and Subj. The results are shown in Table~\ref{tab_different_tempalte}. We can see that the performance gap across various inference templates is smaller than 1 point and this reflects UDR's stability and transferability across different inference templates.
\section{Potential Risk}
Previous works have shown Large language models can have various kinds of bias~\citep{language_model_bias}. Since UDR is trained from the feedback of large language models, it can also contain such bias.

\begin{table*}[t]
\footnotesize
\centering
\begin{tabular}{@{}cccc@{}}
\toprule
Task Family                                        & Task             & Input                            & Output                 \\ \midrule
\multirow{6}{*}{\textit{Sentiment Classification}} & SST-2            & Short Movie Review               & Sentiment Label        \\
& SST-5            & Short Movie Review               & Sentiment Label        \\
& Amazon           & Amazon Product Review            & Sentiment Label        \\
& Yelp             & Yelp Review                      & Sentiment Label        \\
& MR               & Movie Review                     & Sentiment Label        \\
& CR               & Electronics Review               & Sentiment Label        \\ \midrule
\multirow{4}{*}{\textit{Topic Classification}}     & AGNews           & News Article                     & Topic Label            \\
& TREC             & Question                         & Topic Label            \\
& DBPedia          & Wikipedia Text                   & Topic Label            \\
& Yahoo            & Question-answer Pair             & Topic Label            \\ \midrule
\multirow{4}{*}{\textit{Multi-Choice}}             & COPA             & Causal Reasoning Question        & Effect/Cause           \\
& Cosmos QA        & Causal Reasoning Question        & Effect/Cause         \\
& ComV             & Commonsense Hypotheses             & Wrong Hypothesis         \\
& ComE             & Wrong Hypothesis             & Explaination         \\
 \midrule
\multirow{3}{*}{\textit{NLI}}                      
& MNLI             & Image-caption Sentence Pair      & Entailment Label       \\
& SNLI             & Cross-genre Sentence Pair        & Entailment Label       \\
& RTE              & Wikipedia/News Sentence Pair     & Entailment Label       \\ \midrule
\textit{Subjective Classification}                 
& Subj             & Movie Review                     & Subjectivity           \\ \midrule
\textit{Lingustic Acceptibility}                   
& COLA             & Linguistics Publication Sentence & Grammatical Label      \\ \midrule
\multirow{3}{*}{\textit{Semantic Parsing}}         
& BREAK            & Question                         & Question Decomposition \\
& MTOP             & User Utterance                   & TOP Representation     \\
& SMCalFlow        & User Utterance                   & Dataflow Program       \\ \midrule
\multirow{3}{*}{\textit{Text Summarization}}       & CNN/DailyMail    & News Article                     & Highlights             \\
& PubMed           & Scientific Paper's Introduction & Abstract               \\
& Reddit           & Reddit Post                & Summary                \\ \midrule
\textit{Commensense Generation}                    & Commen Gen       & Concepts                  & Coherent Sentence      \\ \midrule
\multirow{2}{*}{\textit{Story Generation}}         & Roc Story        & Head of Story                    & Remaining Story        \\
& Roc Stroy Ending & Four-sentence Story              & Story Ending           \\ \midrule
\multirow{4}{*}{\textit{Code Summarization}}       & Go               & Go Code                          & Documentation          \\
& Python           & Python Code                      & Documentation          \\
& Java             & Java Code                        & Documentation          \\
& PHP              & PHP Code                         & Documentation          \\ \midrule
\multirow{3}{*}{\textit{Text Simplification}}      & WikiAuto         & Wikipedia Sentence               & Simplified Sentence    \\
& WikiAuto-Turk    & Wikipedia Sentence               & Simplified Sentence    \\
& WikiAuto-ASSET   & Wikipedia Sentence               & Simplified Sentence    \\ \midrule
\multirow{2}{*}{\textit{Data to Text}}             & DART             & Triple Set                       & Text                   \\
& E2E              & Key-value Pairs                  & Text                   \\ \bottomrule
\end{tabular}
\caption{The Input/Output Domains of Tasks.}
\label{tab_task_intro}
\end{table*}
\begin{table*}[t]
\footnotesize
\centering
\begin{tabular}{@{}ccccccc@{}}
\toprule
Task Family                                        & Task             & Train  & Dev   & Test  & Report Split & Metric  \\ \midrule
\multirow{6}{*}{\textit{Sentiment Classification}} & SST-2            & 6911   & 873   & 1821  & Test         & Acc     \\
                                                   & SST-5            & 8534   & 1101  & 2210  & Test         & Acc     \\
                                                   & Amazon           & 30000  & 5000  & 3000  & Test         & Acc     \\
                                                   & Yelp             & 30000  & -     & 3000  & Test         & Acc     \\
                                                   & MR               & 8662   & -     & 2000  & Test         & Acc     \\
                                                   & CR               & 1772   & -     & 1996  & Test         & Acc     \\ \midrule
\multirow{4}{*}{\textit{Topic Classification}}     & AGNews           & 29914  & -     & 3000  & Test         & Acc     \\
                                                   & TREC             & 5381   & -     & 500   & Test         & Acc     \\
                                                   & DBPedia          & 30000  & -     & 3000  & Test         & Acc     \\
                                                   & Yahoo            & 29150  & -     & 3000  & Test         & Acc     \\ \midrule
\multirow{4}{*}{\textit{Multi-Choice}}             & COPA             & 500    & -     & 500   & Test         & Acc     \\
                                                   & Cosmos QA        & 18770  & 2603  & 6030  & Dev          & Acc     \\
                                                   & ComE             & 9996   & 997   & 1000  & Test         & Acc     \\
                                                   & ComV             & 9992   & 997   & 1000  & Test         & Acc     \\ \midrule
\multirow{3}{*}{\textit{NLI}}                      & MNLI             & 263789 & 3000  & 9796  & Dev          & Acc     \\
                                                   & SNLI             & 131062 & 3272  & 3262  & Test         & Acc     \\
                                                   & RTE              & 2490   & 277   & 3000  & Dev          & Acc     \\ \midrule
\textit{Subjective Classification}                 & Subj             & 8000   & -     & 2000  & Test         & Acc     \\ \midrule
\textit{Lingustic Acceptibility}                   & COLA             & 8532   & -     & 527   & Test         & Acc     \\ \midrule
\multirow{3}{*}{\textit{Semantic Parsing}}         & BREAK            & 44321  & 7760  & 8069  & Dev          & LF-EM   \\
                                                   & MTOP             & 15667  & 2235  & 4386  & Test         & EM      \\
                                                   & SMCalFlow        & 133584 & 14751 & 22012 & Dev          & EM      \\ \midrule
\multirow{3}{*}{\textit{Text Summarization}}       & CNN/DailyMail    & 155098 & 7512  & 6379  & Test         & Rouge-L \\
& PubMed           & 56254  & 3187  & 3481  & Test         & Rouge-L \\
& Reddit           & 37643  & 576   & 562   & Test         & Rouge-L \\ \midrule
\textit{Commensense Generation}                    & Commen Gen       & 67389  & 993   & 1497  & Dev          & BLEU-3  \\ \midrule
\multirow{2}{*}{\textit{Story Generation}}         & Roc Story        & 87526  & 9799  & 9799  & Test         & BLEU-1  \\
& Roc Stroy Ending & 87906  & 9807  & 9807  & Test         & BLEU-1  \\ \midrule
\multirow{4}{*}{\textit{Code Summarization}}       & Go               & 167137 & 7320  & 8115  & Test         & BLEU-1  \\
& Python           & 250818 & 13841 & 14840 & Test         & BLEU-1  \\
& Java             & 164514 & 5172  & 10928 & Test         & BLEU-1  \\
& PHP              & 240851 & 12964 & 13998 & Test         & BLEU-1  \\ \midrule
\multirow{3}{*}{\textit{Text Simplification}}      & WikiAuto         & 481018 & 1999  & 403   & Test         & SARI    \\
& WikiAuto-Turk    & -      & 1999  & 359   & Test         & SARI    \\
& WikiAuto-ASSET   & -      & 1999  & 359   & Test         & SARI    \\ \midrule
\multirow{2}{*}{\textit{Data to Text}}             & DART             & 30123  & 2718  & 4159  & Test         & BLEU-4  \\
& E2E              & 12563  & 1483  & 1847  & Test         & BLEU-4  \\ \bottomrule
\end{tabular}
\caption{The statistics, split and evaluation metrics of each dataset.}
\label{tab_statistics}
\end{table*}
\clearpage
\onecolumn
\begin{center}
\begin{small}
\begin{xtabular*}{0.95\textwidth}{p{0.92\textwidth}}

        \toprule
            {\em Task Family: Sentiment Classification} \\
            \midrule
            \textbf{Task}: SST-2 \\
            \textbf{Task Instruction}: Sentiment of the sentence: \\
            \textbf{Inference Verbalizer}: \{great, terrible\}\\
            \textbf{Inference Template}: \\
            \textbf{Input:}\\
            A three-hour cinema master class.\\
            \textit{It was terrible.}\\
            A pretensions -- and disposable story --- sink the movie.\\
            \textit{It was great.}\\
            $\cdots$ \\
            The movie 's blatant derivativeness is one reason it 's so lackluster. \\
            \textit{It was} \\
            \textbf{Output:} \\
            \textit{terrible.} \\
        \midrule
        \textbf{Task}: SST-5 \\
        \textbf{Task Instruction}: Sentiment of the sentence: \\
        \textbf{Inference Verbalizer}: \{great, good, okay, bad, terrible\}\\
        \textbf{Inference Template}: Same as SST-2 \\
        \midrule
        \textbf{Task}: Amazon \\
        \textbf{Task Instruction}: Sentiment of the sentence: \\
        \textbf{Inference Verbalizer}: \{great, good, okay, bad, terrible\}\\
        \textbf{Inference Template}: Same as SST-2 \\
        \midrule
        \textbf{Task}: Yelp \\
        \textbf{Task Instruction}: Sentiment of the sentence: \\
        \textbf{Inference Verbalizer}: \{great, good, okay, bad, terrible\}\\
        \textbf{Inference Template}: Same as SST-2 \\
        \midrule
        \textbf{Task}: MR \\
        \textbf{Task Instruction}: Sentiment of the sentence: \\
        \textbf{Inference Verbalizer}: \{great, terrible\}\\
        \textbf{Inference Template}: Same as SST-2 \\
        \midrule
        \textbf{Task}: CR \\
        \textbf{Task Instruction}: Sentiment of the sentence: \\
        \textbf{Inference Verbalizer}: \{great, terrible\}\\
        \textbf{Inference Template}: Same as SST-2 \\
    \midrule
            {\em Task Family: Topic Classification} \\
            \midrule
            \textbf{Task}: AGNews \\
            \textbf{Task Instruction}: Topic of the text: \\
            \textbf{Inference Verbalizer}: \{World, Sports, Business, Technology\}\\
            \textbf{Inference Template}: \\
            \textbf{Input:}\\
            'LONDON, Oct 26 (AFP) - World oil prices will be driven down over the next two years due to there being enough crude to meet soaring demand, Claude Mandil, executive director of the International Energy Agency (IEA), said here Tuesday.\\
            \textit{Topic: Business.}\\
            WASHINGTON - This year's surge in energy prices is likely to have far less of an impact on the economy than the oil shocks of the 1970s, Federal Reserve Chairman Alan Greenspan said Friday.    Greenspan predicted that the global economy will adjust to the recent surge in prices, which has seen oil topping \textbackslash{}\textbackslash{}\$50 per barrel, by boosting energy exploration and production and by increasing fuel efficiency...\\
            \textit{Topic: World.}\\
            $\cdots$ \\
            Oil demand is rising faster than predicted this year as OPEC pumps more low-quality oil in a failed bid to reduce record prices, according to International Energy Agency, an adviser to 26 industrialized nations. \\
            \textit{Topic: } \\
            \textbf{Output:} \\
            \textit{Business.} \\
        \midrule
        \textbf{Task}: TREC \\
        \textbf{Task Instruction}: Topic of the question: \\
        \textbf{Inference Verbalizer}: \{Description, Entity, Expression, Human, Location, Number\}\\
        \textbf{Inference Template}: Same as AGNews \\
        \midrule
        \textbf{Task}: DBPedia \\
        \textbf{Task Instruction}: Topic of the text: \\
        \textbf{Inference Verbalizer}: \{Company, Educational Institution, Artist, Athlete, Office Holder, Mean of Transportation, Building, Natural Place, Village, Animal, Plant, Album, Film, Written Work\}\\
        \textbf{Inference Template}: Same as AGNews \\
        \midrule
        \textbf{Task}: Yahoo \\
        \textbf{Task Instruction}: Topic of the text: \\
        \textbf{Inference Verbalizer}: \{Society \& Culture, Science \& Mathematics, Health, Education \& Reference, Computers \& Internet, Sports, Business \& Finance, Entertainment \& Music, Family \& Relationships, Politics \& Government\}\\
        \textbf{Inference Template}: Same as AGNews \\
    \midrule
            {\em Task Family: Multi-Choice} \\
            \midrule
            \textbf{Task}: COPA \\
            \textbf{Task Instruction}: Answer the question based on the text. \\
            \textbf{Inference Template}: \\
            \textbf{Input:}\\
            I scratched my skin. What happened as a result?\\
            \textit{My itch went away.}\\
             misplaced my wallet. What happened as a result?\\
            \textit{I retraced my steps.}\\
            $\cdots$ \\
            I emptied my pockets. What happened as a result?\\
            \textbf{Output:} \\
            \textit{I retrieved a ticket stub.} \\
        \midrule
        \textbf{Task}: Cosmos QA \\
        \textbf{Task Instruction}: Answer the question based on the text. \\
        \textbf{Inference Template}: Same as COPA \\
        \midrule
        \textbf{Task}: ComV \\
        \textbf{Task Instruction}: Which statement of the two is against common sense? \\
        \textbf{Inference Template}: Same as COPA \\
        \midrule
        \textbf{Task}: ComE \\
        \textbf{Task Instruction}: Select the most corresponding reason why this statement is against common sense. \\
        \textbf{Inference Template}: Same as COPA \\
        \midrule
            {\em Task Family: NLI} \\
            \midrule
            \textbf{Task}: MNLI \\
            \textbf{Task Instruction}: Recognizing textual entailment between these 2 texts. \\
            \textbf{Inference Verbalizer}: \{Entailment, Inconclusive, Contradiction\}\\
            \textbf{Inference Template}: \\
            \textbf{Input:}\\
            uh-huh exactly not what color you are how old you are what if your male or female that would be wonderful i guess it's kind of an ideal world though huh \textit{Based on that information, is the claim} The world would be better if race and gender did not matter. People would get along much better "Entailment", "Contradiction", or "Inconclusive"?\\
            \textit{Answer: Inconclusive.}\\
            uh-huh exactly not what color you are how old you are what if your male or female that would be wonderful i guess it's kind of an ideal world though huh \textit{Based on that information, is the claim} The world would be better if race and gender did not matter.  "Entailment", "Contradiction", or "Inconclusive"?\\
            \textit{Answer: Entailment.}\\
            $\cdots$ \\
            It's that kind of world. \textit{Based on that information, is the claim} The world is getting better. "Entailment", "Contradiction", or "Inconclusive"? \\
            \textit{Answer:} \\
            \textbf{Output:} \\
            \textit{Inconclusive} \\
        \midrule
        \textbf{Task}: SNLI \\
        \textbf{Task Instruction}: Recognizing textual entailment between these 2 texts. \\
        \textbf{Inference Verbalizer}: \{Entailment, Inconclusive, Contradiction\}\\
        \textbf{Inference Template}: Same as MNLI \\
        \midrule
        \textbf{Task}: RTE \\
        \textbf{Task Instruction}: Recognizing textual entailment between these 2 texts. \\
        \textbf{Inference Verbalizer}: \{True, False\}\\
        \textbf{Inference Template}: Same as MNLI \\
    \midrule
            {\em Task Family: Subjective Classification} \\
            \midrule
            \textbf{Task}: Subj \\
            \textbf{Task Instruction}: Subjectivity of the sentence: \\
            \textbf{Inference Verbalizer}: \{subjective, objective\}\\
            \textbf{Inference Template}: \\
            \textbf{Input:}\\
            thirteen conversations about one thing lays out a narrative puzzle that interweaves individual stories , and , like a mobius strip , elliptically loops back to where it began .\\
            \textit{It's subjective.}\\
            a small gem of a movie that defies classification and is as thought-provoking as it is funny , scary and sad .\\
            \textit{It's subjective.}\\
            $\cdots$ \\
            smart and alert , thirteen conversations about one thing is a small gem . \\
            \textit{It's} \\
            \textbf{Output:} \\
            \textit{subjective} \\
        \midrule
            {\em Task Family: Lingustic Acceptibility} \\
            \midrule
            \textbf{Task}: COLA \\
            \textbf{Task Instruction}: The grammaticality of this sentence: \\
            \textbf{Inference Verbalizer}: \{not grammatical, grammatical\}\\
            \textbf{Inference Template}: \\
            \textbf{Input:}\\
            The sea monster drowned the sailors.\\
            \textit{It is grammatical.}\\
            He rode out the storm.\\
            \textit{It is grammatical.}\\
            $\cdots$ \\
            The sailors rode the breeze clear of the rocks. \\
            \textit{It is} \\
            \textbf{Output:} \\
            \textit{grammatical} \\
        \midrule
            {\em Task Family: Semantic Parsing} \\
            \midrule
            \textbf{Task}: BREAK \\
            \textbf{Task Instruction}: Parse the sentence into logical form: \\
            \textbf{Inference Template}: \\
            \textbf{Input:}\\
            \textit{Parse the sentence into logical form:} what flights are available from pittsburgh to boston on saturday\\
            \textit{1\#) return flights 2\#) return \#1 from pittsburgh 3\#) return \#2 to  boston 4\#) return \#3 on saturday 5\#) return \#4 that  are available}\\
            P\textit{arse the sentence into logical form:} what flights are available wednesday afternoon from denver to san francisco\\
            \textit{1\#) return flights 2\#) return \#1 from  denver 3\#) return \#2 to san francisco 4\#) return \#3 on  wednesday afternoon 5\#) return \#4 that  are available}\\
            $\cdots$ \\
            \textit{Parse the sentence into logical form:} what flights are available tomorrow from denver to philadelphia  \\
            \textit{1\#)} \\
            \textbf{Output:} \\
            \textit{return flights ;return \#1 from  denver ;return \#2 to philadelphia ;return \#3 if  available} \\
        \midrule
        \textbf{Task}: MTOP \\
        \textbf{Task Instruction}: Parse the sentence into logical form: \\
        \textbf{Inference Template}: Same as BREAK \\
        \midrule
        \textbf{Task}: SMCalFlow \\
        \textbf{Task Instruction}: Parse the sentence into logical form: \\
        \textbf{Inference Template}: Same as BREAK \\
        \midrule
            {\em Task Family: Text Summarization} \\
            \midrule
            \textbf{Task}: CNN/DailyMail \\
            \textbf{Task Instruction}: Summarize the text: \\
            \textbf{Inference Template}: \\
            \textbf{Input:}\\
            \textit{Summarize the text:} JERUSALEM (CNN) -- Israel moved to defend itself in the face of international criticism Monday over its eviction of dozens of Palestinian families from a neighborhood of Jerusalem they have lived in for generations.$\cdots$\\
            \textit{TL;DR: Israel incurs international criticism over eviction of Palestinian families .
Two Jewish families moved in after evictions in East Jerusalem .
Israeli spokesman says dispute is a legal one between private parties .}\\
            \textit{Summarize the text:} (CNN)The International Criminal Court opened an inquiry into attacks in Palestinian territories, paving the way for possible war crimes investigation against Israelis.$\cdots$\\
            \textit{TL;DR: An inquiry allows the court to review evidence and determine whether to file charges .
The U.S. calls for negotiations between Palestinian, Israeli officials .}\\
            $\cdots$ \\
            \textit{Summarize the text:} (CNN)The Palestinian Authority officially became the 123rd member of the International Criminal Court on Wednesday, a step that gives the court jurisdiction over alleged crimes in Palestinian territories.$\cdots$ \\
            \textit{TL;DR:} \\
            \textbf{Output:} \\
            \textit{Membership gives the ICC jurisdiction over alleged crimes committed in Palestinian territories since last June .
Israel and the United States opposed the move, which could open the door to war crimes investigations against Israelis .} \\
        \midrule
        \textbf{Task}: PubMed \\
        \textbf{Task Instruction}: Summarize the text: \\
        \textbf{Inference Template}: Same as CNN/DailyMail \\
        \midrule
        \textbf{Task}: Reddit \\
        \textbf{Task Instruction}: Summarize the text: \\
        \textbf{Inference Template}: Same as CNN/DailyMail \\
        \midrule
            {\em Task Family: Commensense Generation} \\
            \midrule
            \textbf{Task}: Commen Gen \\
            \textbf{Task Instruction}: Generate a sentence using these concepts: \\
            \textbf{Inference Template}: \\
            \textbf{Input:}\\
            Generate a sentence using these concepts: counter, pizza, restaurant\\
            \textit{Generated sentence: Two men standing at counters assembling pizzas in a restaurant.}\\
            Generate a sentence using these concepts: counter, restaurant, stand\\
            \textit{Generated sentence: A man stands behind the counter of a restaurant.}\\
            $\cdots$ \\
            Generate a sentence using these concepts: field, look, stand \\
            \textit{Generated sentence:} \\
            \textbf{Output:} \\
            \textit{The player stood in the field looking at the batter.} \\
        \midrule
            {\em Task Family: Story Generation} \\
            \midrule
            \textbf{Task}: Roc Story \\
            \textbf{Task Instruction}: Beginning of the story: \\
            \textbf{Inference Template}: \\
            \textbf{Input:}\\
            \textit{Beginning of the story:} Taylor had been up all night memorizing lines for the play.\\
            \textit{Rest of the story: She knew that most of the girls in her class would be auditioning too. She watched the other girls stumble over their lines. She took a deep breath before going up on stage with a smile. All her lines were delivered perfectly and she got the part.}\\
            \textit{Beginning of the story:} Gabby was proud to be given the lead role in the school play.\\
            \textit{Rest of the story: She had worked hard on her audition piece. She worked hard to memorize her lines for the play. When the show opened, she stood on the stage and took it all in. She loved the feeling of performing.}\\
            $\cdots$ \\
            \textit{Beginning of the story:} Natalie had auditioned for the lead in the school play. \\
            \textit{Rest of the story:} \\
            \textbf{Output:} \\
            \textit{She won the part and was super excited. She rehearsed for weeks and weeks. On opening night, she acted her little heart out. The play was a huge success!} \\
        \midrule
        \textbf{Task}: Roc Story Ending \\
        \textbf{Task Instruction}: An unfinished story: \\
        \textbf{Inference Template}: Same as ROC Story \\
        \midrule
            {\em Task Family: Code Summarization} \\
            \midrule
            \textbf{Task}: Go \\
            \textbf{Task Instruction}: Comment on the code. \\
            \textbf{Inference Template}: \\
            \textbf{Input:}\\
            \textit{Comment on the code. Code:} \\
func NewSTM ( c * v3 . Client , apply func ( STM ) error , so ... stmOption ) ( * v3 . TxnResponse , error ) \{ \\
\ \ \ \ opts :=  \& stmOptions  \{ \\
\ \ \ \ \ \ \ \ ctx : c . Ctx ( )   \\
\ \ \ \ \} \\
\ \ \ \ for  \_ , f := range so  \{ \\
\ \ \ \ \ \ \ \ f ( opts )  \\
\ \ \ \ \}  \\
\ \ \ \ if len ( opts . prefetch ) != 0  \{  \\
\ \ \ \ \ \ \ \ f := apply apply = func ( s STM ) error  \{  \\
\ \ \ \ \ \ \ \ \ \ \ \ s . Get ( opts . prefetch ... )  \\
\ \ \ \ \ \ \ \ \ \ \ \ return f ( s )   \\
\ \ \ \ \ \ \ \ \}   \\
\ \ \ \ \}  \\
\ \ \ \ return runSTM ( mkSTM ( c , opts ) , apply )   \\
\} \\\\
            \textit{Comment: RunContainer runs a fake Docker container}\\
            \textit{Comment on the code. Code:} \\
            func ( s \* Subnet ) EnsureDead ( ) ( err error ) \{ \\
\ \ \ \ defer errors . DeferredAnnotatef ( \& err , " " , s ) \\
\ \ \ \ if s . doc . Life == Dead \{ \\
\ \ \ \ \ \ \ \ return nil \\
\ \ \ \ \} \\
\ \ \ \ ops := [ ] txn . Op \{ \{ \\
\ \ \ \ \ \ \ \ \ \ \ \ C : subnetsC , Id : s . doc . DocID , Update : bson . D \{ \{ " " , bson . D \{ \{ " " , Dead \} \} \} \} , Assert : isAliveDoc , \\
\ \ \ \ \ \ \ \ \} \} \\
\ \ \ \ txnErr := s . st . db ( ) . RunTransaction ( ops ) \\
\ \ \ \ if txnErr == nil \{ \\
\ \ \ \ \ \ \ \ s . doc . Life = Dead return nil \\
\ \ \ \ \} \\
\ \ \ \ return onAbort ( txnErr , subnetNotAliveErr ) 
\}\\
            \textit{Comment: EnsureDead sets the Life of the subnet to Dead if it s Alive . If the subnet is already Dead no error is returned . When the subnet is no longer Alive or already removed errNotAlive is returned .}\\
            $\cdots$ \\
            \textit{Comment on the code. Code:} \\
            func NewSTM ( c * v3 . Client , apply func ( STM ) error , so ... stmOption ) ( * v3 . TxnResponse , error ) \{ \\
\ \ \ \ opts := \& stmOptions \{ ctx : c . Ctx ( ) \} \\
\ \ \ \ for \_ , f := range so \{ f ( opts ) \} \\
\ \ \ \ if len ( opts . prefetch ) != 0 \{ \\
\ \ \ \ \ \ \ \ f := apply apply = func ( s STM ) error \{ s . Get ( opts . prefetch ... ) return f ( s ) \} \\
\ \ \ \ \} \\
\ \ \ \ return runSTM ( mkSTM ( c , opts ) , apply ) \\
\} \\
            \textit{Comment:} \\
            \textbf{Output:} \\
            \textit{NewSTM initiates a new STM instance using serializable snapshot isolation by default .} \\
        \midrule
        \textbf{Task}: Python \\
        \textbf{Task Instruction}: Comment on the code. \\
        \textbf{Inference Template}: Same as Go \\
        \midrule
        \textbf{Task}: Java \\
        \textbf{Task Instruction}: Comment on the code. \\
        \textbf{Inference Template}: Same as Go \\
        \midrule
        \textbf{Task}: PHP \\
        \textbf{Task Instruction}: Comment on the code. \\
        \textbf{Inference Template}: Same as Go \\

        \midrule
            {\em Task Family: Text Simplification} \\
            \midrule
            \textbf{Task}: WikiAuto \\
            \textbf{Task Instruction}: Simplify the text: \\
            \textbf{Inference Template}: \\
            \textbf{Input:}\\
            \textit{Simplify the text:} Stanton went on to write some of the most influential books , documents , and speeches of the women 's rights movement .\\
            \textit{Simplified text: Together they wrote speeches , articles , and books .}\\
            \textit{Simplify the text:} When she was eighteen and without a university education , she began writing for the newspaper " Exce  lsior " , doing interviews and society columns .\\
            \textit{Simplified text: She began writing for the newspaper " Exce  lsior " , doing interviews and society columns .}\\
            $\cdots$ \\
            \textit{Simplify the text:} Together with James, she compiled crosswords for several newspapers and magazines, including People, and it was in 1978 that they launched their own publishing company. \\
            \textit{Simplified text:} \\
            \textbf{Output:} \\
            \textit{Together with James, she compiled crosswords. It was for several newspapers and magazines, including People. They launched their own publishing company. It was in 1978.} \\
        \midrule
        \textbf{Task}: WikiAuto-Turk \\
        \textbf{Task Instruction}: Simplify the text: \\
        \textbf{Inference Template}: Same as WikiAuto \\
        \midrule
        \textbf{Task}: WikiAuto-ASSET \\
        \textbf{Task Instruction}: Simplify the text: \\
        \textbf{Inference Template}: Same as WikiAuto \\
        \midrule
            {\em Task Family: Data to Text} \\
            \midrule
            \textbf{Task}: DART \\
            \textbf{Task Instruction}: Describe the table in natural language. \\
            \textbf{Inference Template}: \\
            \textbf{Input:}\\
            \textit{Describe the table in natural language. Table:} [Baywatch | NOTES | Episode: Red Wind], [Baywatch | ROLE | Kim], [[TABLECONTEXT] | TITLE | Baywatch], [[TABLECONTEXT] | [TITLE] | Bobbie Phillips]\\
            \textit{Sentence: Bobbie Phillips appeared on the episode Red Wind in Baywatch as Kim.}\\
            \textit{Describe the table in natural language. Table:} [Silk Stalkings | ROLE | Tessa Shaver], [[TABLECONTEXT] | [TITLE] | Bobbie Phillips], [[TABLECONTEXT] | TITLE | Silk Stalkings], [Silk Stalkings | NOTES | Episode: Goodtime Charlie]\\
            \textit{Sentence: Actress Bobbie Phillips was casted as Tessa Shaver on the episode Goodtime Charlie of Silk Stalkings.}\\
            $\cdots$ \\
            \textit{Describe the table in natural language. Table:} [Hawaii Five-O | NOTES | Episode: The Flight of the Jewels], [[TABLECONTEXT] | [TITLE] | Jeff Daniels], [[TABLECONTEXT] | TITLE | Hawaii Five-O] \\
            \textit{Sentence: } \\
            \textbf{Output:} \\
            \textit{Jeff Daniels played in the Hawaii Five-O episode The Flight of the Jewels} \\
        \midrule
        \textbf{Task}: E2E \\
        \textbf{Task Instruction}: Describe the table in natural language. \\
        \textbf{Inference Template}: Same as DART \\
        \bottomrule

\end{xtabular*}
\captionof{table}{The instructions, inference templates and example cases of tasks.}
\label{tab_templates}
\end{small}
\end{center}
\twocolumn

\end{document}